\def\ThisFile{DataQualityViaDataDegradation.tex}
\def\ThisFileDate{2021/12/24}
\documentclass[11pt]{article}
\usepackage{ifthen,pifont,latexsym}
\usepackage[T1]{fontenc}
\usepackage{color}
\usepackage{times,avant}
\usepackage[round,authoryear]{natbib}
\usepackage{amsmath}
\usepackage{setspace}
\usepackage{xcolor,colortbl}

\usepackage[pdftex]{graphicx}
\definecolor{red}{rgb}{1,.5,.5}

{\end{description}}
{\end{itemize}}
{\end{enumerate}}
\def\Includefigs{true} 
\def\Thisfile{acronyms.tex}
\def\Thisfiledate{2021/08/13}
\def\NISS{National Institute of Statistical Sciences (NISS)
    \gdef\NISS{NISS}}

\def\NCSU{North Carolina State University (NCSU)
    \gdef\NCSU{NCSU}}
\def\UNC{University of North Carolina at Chapel Hill (UNC)
    \gdef\UNC{UNC}}
\def\CMU{Carnegie Mellon University (CMU)
    \gdef\CMU{CMU}}
\def\DU{Duke University (Duke)
    \gdef\DU{Duke}}
\def\UMI{University of Michigan (UMI)
    \gdef\UMI{UMi}}
\def\UMD{University of Maryland College Park (UMD)
    \gdef\UMD{UMd}}
\def\PU{Purdue University (Purdue)
    \gdef\PU{Purdue}}
\def\SMU{Southern Methodist University (SMU)
    \gdef\SMU{SMU}}
\def\GMU{George Mason University (GMU)
    \gdef\GMU{GMU}}
\def\UIC{University of Illinois at Chicago (UIC)
    \gdef\UIC{UIC}}
\def\LANL{Los Alamos National Laboratory (LANL)
    \gdef\LANL{LANL}}
\def\PNNL{Pacific Northwest National Laboratory (PNNL)
    \gdef\PNNL{PNNL}}
\def\GM{General Motors (GM)
    \gdef\GM{GM}}
\def\GSK{GlaxoSmithKline (GSK)
    \gdef\GSK{GSK}}
\def\VI{Visual Insights (VI)
    \gdef\VI{VI}}
\def\EIA{Energy Information Administration (EIA)
    \gdef\EIA{EIA}}
\def\EPA{Environmental Protection Agency (EPA)
    \gdef\EPA{EPA}}
\def\NCES{National Center for Education Statistics (NCES)
    \gdef\NCES{NCES}}
\def\BTS{Bureau of Transportation Statistics (BTS)
    \gdef\BTS{BTS}}
\def\BLS{Bureau of Labor Statistics (BLS)
    \gdef\BLS{BLS}}
\def\NCHS{National Center for Health Statistics (NCHS)
    \gdef\NCHS{NCHS}}
\def\BC{Census Bureau (Census)
    \gdef\BC{Census}}
\def\CB{Census Bureau (Census)
    \gdef\CB{Census}}
\def\NASS{National Agricultural Statistics Service (NASS)
    \gdef\NASS{NASS}}
\def\NSF{National Science Foundation (NSF)
    \gdef\NSF{NSF}}
\def\DMS{Division of Mathematical Sciences (DMS)
    \gdef\DMS{DMS}}
\def\CISE{Computer and Information Sciences and Engineering 
    (CISE)\gdef\CISE{CISE}}
\def\CATS{Committee on Theoretical and Applied Statistics (CATS)
    \gdef\CATS{CATS}}
\def\NRC{National Research Council (NRC)
    \gdef\NRC{NRC}}
\def\CNSTAT{Committee on National Statistics (CNSTAT)
    \gdef\CNSTAT{CNSTAT}}
\def\DOD{US Department of Defense (DoD)
    \gdef\DOD{DoD}}
\def\USGS{US Geological Survey (USGS)
    \gdef\USGS{USGS}}
\def\OMB{Office of Management and Budget (OMB)
    \gdef\OMB{OMB}}
\def\NSA{National Security Agency (NSA)
    \gdef\NSA{NSA}}
\def\DHS{Department of Homeland Security (DHS)
    \gdef\DHS{DHS}}
\def\CDC{Centers for Disease Control and Prevention (CDC)
    \gdef\CDC{CDC}}
\def\DARPA{Defense Advanced Research Projects Agency (DARPA)
    \gdef\DARPA{DARPA}}
\def\DOE{Department of Energy (DOE)
    \gdef\DOE{DOE}}
\def\CIA{Central Intelligence Agency (CIA)
    \gdef\CIA{CIA}}
\def\DTRA{Defense Threat Reduction Agency (DTRA)
    \gdef\DTRA{DTRA}}
\def\NIST{National Institute of Standards and Technology (NIST)
    \gdef\NIST{NIST}}
\def\NIAAA{National Institute on Alcohol Abuse and Alcoholism
    (NIAAA)
    \gdef\NIAAA{NIAAA}}
\def\ARO{Army Research Office (ARO)
    \gdef\ARO{ARO}}
\def\FDA{Food and Drug Administration (FDA)
    \gdef\FDA{FDA}}
\def\SAMSI{Statistical and Applied Mathematical Sciences
    Institute (SAMSI)\gdef\SAMSI{SAMSI}}

\def\NCDOT{North Carolina Department of Transporation (NCDOT)
    \gdef\NCDOT{NCDOT}}
\def\NCGBC{North Carolina Bioinformatics and Genomics Consortium 
    (NCGBC)\gdef\NCGBC{NCGBC}}
\def\RTI{RTI International (RTI)
    \gdef\RTI{RTI}}
\def\CIIT{CIIT Centers for Health Research (CIIT)
    \gdef\CIIT{CIIT}}
\def\DQRI{Data Quality Research Institute (DQRI)
    \gdef\DQRI{DQRI}}
\def\DIMACS{Center for Discrete Mathematics and Theoretical %
    Computer Science (DIMACS)\gdef\DIMACS{DIMACS}}
\def\HDF{Hereditary Disease Foundation (HDF)
    \gdef\HDF{HDF}}
\def\NCDM{National Center for Data Mining (NCDM)
    \gdef\NCDM{NCDM}}
\def\RTP{Research Triangle Park (RTP)
    \gdef\RTP{RTP}}
\def\ITDB{Intermodal Transportation Database (ITDB)
    \gdef\ITDB{ITDB}}
\def\TRI{Toxic Release Inventory (TRI)
    \gdef\TRI{TRI}}
\def\CPS{Current Population Survey (CPS)
    \gdef\CPS{CPS}}
\def\SASS{Schools and Staffing Survey (SASS)
    \gdef\SASS{SASS}}
\def\ITR{Information Technology Research (ITR)
    \gdef\ITR{ITR}}
\def\DC{data confidentiality (DC)
    \gdef\DC{DC}}
\def\DQ{data quality (DQ)
    \gdef\DQ{DQ}}
\def\DI{data integration (DI)
    \gdef\DI{DI}}
\def\IT{information technology (IT)
    \gdef\IT{IT}}
\def\SDL{statistical disclosure limitation (SDL)
    \gdef\SDL{SDL}}
\def\IQ{information quality (IQ)
    \gdef\IQ{IQ}}
\def\MCMC{Markov chain Monte Carlo (MCMC)
    \gdef\MCMC{MCMC}}
\def\CSV{comma-separated value (CSV)
    \gdef\CSV{CSV}}

\def\RMI{remote method invocation (RMI)
    \gdef\RMI{RMI}}
\def\SOAP{simple object access protocol (SOAP)
    \gdef\SOAP{SOAP}}
\def\XML{extensible markup language (XML)
    \gdef\XML{XML}}

\def\NHTSA{National Highway Traffic Safety Administration (NHTSA)%
    \gdef\NHTSA{NHTSA}}
\def\RDB{relational database (RDB)
    \gdef\RDB{RDB}\gdef\RDBMS{RBDMS}\gdef\RDBMSs{RDBMSs}}
\def\RDBMS{relational database management system (RDBMS)
    \gdef\RDB{RDB}\gdef\RDBMS{RBDMS}\gdef\RDBMSs{RDBMSs}}
\def\RDBMSs{relational database management systems (RDBMSs)
    \gdef\RDB{RDB}\gdef\RDBMS{RBDMS}\gdef\RDBMSs{RDBMSs}}
\def\GIS{geographical information system (GIS)\gdef\GIS{GIS}}
\def\DQTK{data quality toolkit (DQTK)
    \gdef\DQTK{DQTK}}
\def\DQRC{data quality report card (DQRC)
    \gdef\DQRC{DQRC}}
\def\TDQM{Total Data Quality Management (TQDM)
    \gdef\TDQM{TDQM}}
\def\OTR{optimal tabular release (OTR)
    \gdef\OTR{OTR}\gdef\OTRs{OTRs}}
\def\OTRs{optimal tabular releases (OTRs)
    \gdef\OTR{OTR}\gdef\OTRs{OTRs}}
\def\GUI{graphical user interface (GUI)
    \gdef\GUI{GUI}}
\def\OLTP{On-Line Transaction Processing (OLTP)
    \gdef\OLTP{OLTP}}
\def\GPRA{Government Performance Results Act (GPRA)
    \gdef\GPRA{GPRA}}

\def\CRM{Customer Relationship Management (CRM)
    \gdef\CRM{CRM}}
\def\HCI{human--computer interaction (HCI)
    \gdef\HCI{HCI}}
\def\NDHS{National Defense and Homeland Security (NDHS)
    \gdef\NDHS{NDHS}}
\def\MMR04{2004 Conference on Mathematical Methods in Reliability (MMR 2004)
    \gdef\MMR04{MMR 2004}}

\def\NCLB{No Child Left Behind Act (NCLB)
    \gdef\NCLB{NCLB}}
\def\CCD{Common Core of Data (CCD)
    \gdef\CCD{CCD}}

\def\NIH{National Institutes of Health (NIH)
    \gdef\NIH{NIH}}
\def\EFF{Electronic Frontier Foundation (EFF)
    \gdef\EFF{EFF}}
\def\GCD{graduation, completion and dropout (GCD)
    \gdef\GCD{GCD}}

\def\NCAR{National Center for Atmospheric Research (NCAR)
    \gdef\NCAR{NCAR}}

\def\TIMSS{Third International Mathematics and Science Study (TIMSS)
    \gdef\TIMSS{TIMSS}}
\def\PISA{Program for International Student Assessment (PISA)
    \gdef\PISA{PISA}}
\def\IEA{International Association for the Evaluation of
        Educational Achievement (IEA)
    \gdef\IEA{IEA}}
\def\NAEP{National Assessment of Educational Progress (NAEP)
    \gdef\NAEP{NAEP}}
\def\PIRLS{Progress in International Reading Literacy Study (PIRLS)
    \gdef\PIRLS{PIRLS}}

\def\OECD{Organization for Economic Cooperation and Development (OECD)
    \gdef\OECD{OECD}}

\def\GDC{graduation, dropout and completion (GDC)
    \gdef\GDC{GDC}}

\def\SMPC{secure multi-party computation (SMPC)
    \gdef\SMPC{SMPC}}
\def\HIPAA{Health Insurance Privacy and Accountability Act (HIPAA)
    \gdef\HIPAA{HIPAA}}

\def\ACS{American Community Survey (ACS)
    \gdef\ACS{ACS}}

\def\ESSI{Education Statistics Services Institute (ESSI)
    \gdef\ESSI{ESSI}}

\def\SCDM{Society for Clinical Data Management (SCDM)
    \gdef\SCDM{SCDM}}
\def\ACDM{Association for Clinical Data Management (ACDM)
    \gdef\ACDM{ACDM}}

\def\ASA{American Statistical Association (ASA)
    \gdef\ASA{ASA}}

\def\NCAA{National Collegiate Athletic Association (NCAA)
    \gdef\NCAA{NCAA}}
\def\GED{General Education Development (GED)
    \gdef\GED{GED}}

\def\ISU{Iowa State University (ISU)
    \gdef\ISU{ISU}}

\def\FIPS{Federal Information Processing System (FIPS)
    \gdef\FIPS{FIPS}}
\def\GSA{General Services Administration (GSA)
    \gdef\GSA{GSA}}

\def\AIR{American Institutes for Research (AIR)
    \gdef\AIR{AIR}}
\def\ESSIS{Education Statistics Services Institute---Statistics
    (ESSI--Stat)\gdef\ESSIS{ESSI-Stat}}

\def\NESSI{NAEP Education Statistics Services Institute (NESSI)
    \gdef\NESSI{NESSI}}

\def\ACM{Association for Computing Machinery (ACM)
    \gdef\ACM{ACM}}
\def\IEEE{Institute of Electrical and Electronics Engineers (IEEE)
    \gdef\IEEE{IEEE}}
\gdef\SIAM{Society for Industrial and Applied Mathematics (SIAM)
    \gdef\SIAM{SIAM}}
\def\IAOS{ISI Section on Official Statistics (IAOS)
    \gdef\IAOS{IAOS}}
\def\ISBA{International Society for Bayesian Analysis (ISBA)
    \gdef\ISBA{ISBA}}

\def\CDAC{Confidentiality and Data Access Committee (CDAC)
    \gdef\CDAC{CDAC}}
\def\CSIRO{Commonwealth Scientific and Industrial Research
    Organisation (CSIRO)\gdef\CSIRO{CSIRO}}
\def\TUCASI{Triangle Universities Center for Advanced Studies,
    Inc.\ (TUCASI)\gdef\TUCASI{TUCASI}}
\def\NCI{National Cancer Institute (NCI)
    \gdef\NCI{NCI}}

\def\BMSA{Board on Mathematical Sciences and their Applications
    (BMSA)\gdef\BMSA{BMSA}}

\def\NSCAW{National Survey of Child and Adolescent Well-Being
    (NSCAW)\def\NSCAW{NSCAW}}

\def\DAS{Data Analysis System (DAS)
    \gdef\DAS{DAS}\gdef\DASs{DAS's}}
\def\DASs{Data Analysis Systems (DAS's)
    \gdef\DAS{DAS}\gdef\DAS{DAS's}}

\def\CFFR{Committee on Federally Funded Research (CFFR)
    \gdef\CFFR{CFFR}}

\def\AJS{American Judicature Society (AJS)
    \gdef\AJS{AJS}}
\def\ECCR{Exploratory Center for Cheminformatics Research (ECCR)
    \gdef\ECCR{ECCR}}
\def\AAAS{American Association for the Advancement of Science (AAAS)
    \gdef\AAAS{AAAS}}

\def\RTF{Research Triangle Foundation (RTF)
    \gdef\RTF{RTF}}

\def\CFFR{Committee on Federally Funded Research (CFFR)%
    \gdef\CFFR{CFFR}}
\def\MPS{Directorate for Mathematical and Physical Sciences (MPS)%
    \gdef\MPS{MPS}}
\def\CMG{Collaborations in Mathematical Geosciences (CMG)%
    \gdef\CMG{CMG}}

\def\IMS{Institute of Mathematical Statistics (IMS)
    \gdef\IMS{IMS}}
\def\ISI{International Statistical Institute (ISI)
    \def\ISI{ISI}}
\def\IFNA{Interface Foundation of North America (IFNA)
    \gdef\IFNA{IFNA}}

\def\PPDM{privacy-preserving data mining (PPDM)
    \gdef\PPDM{PPDM}}

\def\ITSEW{International Total Survey Error Workshops (ITSEW)
    \gdef\ITSEW{ITSEW}}

\def\STEM{science, technology, engineering and mathematics (STEM)
    \gdef\STEM{STEM}}

\def\ASG{Art and Science Group (A\&SG)
    \gdef\ASG{A\&SG}}

\def\HSLS{High School Longitudinal Study (HSLS:09)
    \gdef\HSLS{HSLS:09}}
\def\NHIS{National Health Interview Survey (NHIS)
    \gdef\NHIS{NHIS}}
\def\FRG{Focused Research Groups in the Mathematical Sciences (FRG)
    \gdef\FRG{FRG}}

\def\IES{Institute of Education Sciences (IES)
    \gdef\IES{IES}}
\def\IOJ{Institute of Justice (IOJ)
    \gdef\IOJ{IOJ}}
\def\EAA{experimental analysis of algorithms (EAA)
    \gdef\EAA{EAA}\gdef\EAAC{EAA}}
\def\EAAC{Experimental analysis of algorithms (EAA)
    \gdef\EAA{EAA}\gdef\EAAC{EAA}}

\def\JPC{\textit{Journal of Privacy and Confidentialty} (JPC)
    \gdef\JPC{\textit{JPC}}}

\def\NWG{NISS Working Group (NWG)
    \gdef\NWG{NWG}\gdef\NWGS{NWGS}}
\def\NWGS{NISS Working Groups (NWGs)
    \gdef\NWGS{NWGS}\gdef\NWG{NWG}}

\def\DHHS{Department of Health and Human Services (DHHS)
    \gdef\DHHS{DHHS}}
\def\OCC{Office of the Comptroller of the Currency (OCC)
    \gdef\OCC{OCC}}

\def\FIPSE{Fund for the Improvement of Postsecondary Education
(FIPSE)
    \gdef\FIPSE{FIPSE}}

\def\FHWA{Federal Highway Administration (FHWA)\gdef\FHWA{FHWA}}

\def\SPAIG{Statistical Partnerships among Academia, Industry and
    Government (SPAIG)\gdef\SPAIG{SPAIG}}
\def\COPSS{Committee of Presidents of Statistical Societies (COPSS)
    \gdef\COPSS{COPSS}}

\def\SDDS{School District Demographics System (SDDS)
    \gdef\SDDS{SDDS}}

\def\ITRE{Institute for Transportation Research and Education (ITRE)
    \gdef\ITRE{ITRE}}
\def\TRB{Transportation Research Board (TRB)
    \gdef\TRB{TRB}}

\def\DTRA{Defense Threat Reduction Agency (DTRA)
    \def\DTRA{DTRA}}

\def\CPTAC{Clinical Proteomic Technology Assessment for Cancer (CPTAC)
    \gdef\CPTAC{CPTAC}}

\def\SRS{Division of Science Resources Statistics (SRS)
    \gdef\SRS{SRS}}
\def\SED{Survey of Earned Doctorates (SED)
    \gdef\SED{SED}}
\def\SDR{Survey of Doctorate Recipients (SDR)
    \gdef\SDR{SDR}}
\def\GSS{Survey of Graduate Students and Postdoctorates in
    Science and Engineering (GSS)
    \gdef\GSS{GSS}}
\def\RCG{National Survey of Recent College Graduates (RCG)
    \gdef\RCG{RCG}}
\def\NSCG{National Survey of College Graduates (NSCG)
    \gdef\NSCG{NCSG}}
\def\BRDIS{Business R\&D Innovation Survey (BRDIS)
    \gdef\BRDIS{BRDIS}}
\def\SANDE{science and engineering (S\&E)
    \gdef\SANDE{S\&E}}
\def\IPEDS{Integrated Postsecondary Education Data System (IPEDS)
    \gdef\IPEDS{IPEDS}}

\def\SEW{science, engineering and health workforce (SEHW)
    \gdef\SEW{SEHW}}

\def\CIPSEA{Confidential Information Protection and Statistical
    Efficiency Act of 2002 (CIPSEA)
    \gdef\CIPSEA{CIPSEA}}

\def\PSU{primary sampling unit (PSU)
    \gdef\PSU{PSU}\gdef\PSUS{PSUs}}
\def\PSUS{primary sampling units (PSUs)
    \gdef\PSU{PSU}\gdef\PSUS{PSUs}}

\def\NHANES{National Health and Nutrition Examination Survey (NHANES)
    \gdef\NHANES{NHANES}}
\def\ECLS{Early Childhood Longitudinal Study (ECLS)
    \gdef\ECLS{ECLS}}

\def\FERPA{Family Educational Rights and Privacy Act (FERPA)
    \gdef\FERPA{FERPA}}

\def\LBD{Longitudinal Business Database (LBD)
    \gdef\LBD{LBD}}
\def\LEHD{Longitudinal Employer-Household Dynamics (LEHD)
    \gdef\LEHD{LEHD}}
\def\SLDS{statewide longitudinal data systems (SLDS)
    \gdef\SLDS{SLDS}}

\def\ECLSK{Early Childhood Longitudinal Study--Kindergarten Class of
    1998--99 (ECLS-K)\gdef\ECLSK{ECLS-K}}
\def\SESTAT{Scientists and Engineers Statistical Data System (SESTAT)
    \gdef\SESTAT{SESTAT}}

\def\RENCI{Renaissance Computing Institute (RENCI)
    \gdef\RENCI{RENCI}}

\def\NIA{National Institute on Aging (NIA)
    \gdef\NIA{NIA}}
\def\SCOPE{Statistical Community of Practice and Engagement (SCOPE)
    \gdef\SCOPE{SCOPE}}
\def\ICSP{Interagency Council on Statistical Policy (ICSP)
    \gdef\ICSP{ICSP}}

\def\WSSM{World's Simplest Survey Microsimulator (WSSM)
    \gdef\WSSM{WSSM}}
\def\CES{Consumer Expenditure Survey (CES)
    \gdef\CES{CES}}

\def\NCSES{National Center for Science and Engineering Statistics (NCSES)
    \gdef\NCSES{NCSES}}

\def\TEP{Technical Expert Panel (TEP)
    \gdef\TEP{TEP}}
\def\ESSIN{Education Statistics Support Institute Network (ESSIN)
    \gdef\ESSIN{ESSIN}}
\def\FCSM{Federal Committee on Statistical Methodology (FCSM)
    \gdef\FCSM{FCSM}}
\def\JSM{Joint Statistical Meetings (JSM)
    \gdef\JSM{JSM}}
\def\NRBA{nonresponse bias analysis (NRBA)
    \gdef\NRBA{NRBA}}
\def\PSS{Private School Survey (PSS)
    \gdef\PSS{PSS}}
\def\CWI{comparable wage index (CWI)
    \gdef\CWI{CWI}}

\def\TSE{total survey error (TSE)
    \gdef\TSE{TSE}}

\def\TCRN{Triangle Census Research Network (TCRN)
    \gdef\TCRN{TCRN}}
\def\CER{comparative effectiveness research (CER)
    \gdef\CER{CER}}

\def\PII{personally identifiable information (PII)
    \gdef\PII{PII}}

\def\OES{Occupational Employment Statistics (OES)
    \gdef\OES{OES}}
\def\CPI{Consumer Price Index (CPI)
    \gdef\CPI{CPI}}
\def\CBSA{Core Based Statistical Area (CBSA)
    \gdef\CBSA{CBSA}\gdef\CBSAS{CBSAs}}
\def\CBSAS{Core Based Statistical Areas (CBSAs)
    \gdef\CBSA{CBSA}\gdef\CBSAS{CBSAs}}
\def\PUMA{Public Use Microdata Area (PUMA)
    \gdef\PUMA{PUMA}\gdef\PUMAS{PUMAs}}
\def\PUMAS{Public Use Microdata Areas (PUMAs)
    \gdef\PUMA{PUMA}\gdef\PUMAS{PUMAs}}
\def\PUMS{public use microdata samples (PUMS)
    \gdef\PUMS{PUMS}}

\def\RDC{Research Data Center (RDC)
    \gdef\RDC{RDC}\gdef\RDCS{RDCs}}
\def\RDCS{Research Data Centers (RDCs)
    \gdef\RDC{RDC}\gdef\RDCS{RDCs}}

\def\NCRN{NSF--Census Research Network (NCRN)
    \gdef\NCRN{NCRN}}
\def\NCRNCO{NSF--Census Research Network Coordination Office (NCRN-CO)
    \gdef\NCRNCO{NCRN-CO}}

\def\MPR{Mathematica Policy Research (MPR)
    \gdef\MPR{MPR}}
\def\NORC{NORC at the University of Chicago (NORC)
    \gdef\NORC{NORC}}
\def\AAPOR{American Association for Public Opinion Research (AAPOR)
    \gdef\AAPOR{AAPOR}}
\def\AEA{American Economic Association (AEA)
    \gdef\AEA{AEA}}

\def\SC{Steering Committee (SC)
    \gdef\SC{SC}}
\def\AEA{American Economic Association (AEA)
    \gdef\AEA{AEA}}
\def\ICES{International Conference on Establishment Statistics (ICES)
    \def\ICES{ICES}}
\def\IRS{Internal Revenue Service (IRS)
    \gdef\IRS{IRS}}
\def\SSA{Social Security Administration (SSA)
    \gdef\SSA{SSA}}
\def\COSSA{Consortium of Social Science Associations (COSSA)
    \gdef\COSSA{COSSA}}
\def\COPAFS{Council of Professional Associations on Federal Statistics (COPAFS)
    \gdef\COPAFS{COPAFS}}
\def\BJS{Bureau of Justice Statistics (BJS)
    \gdef\BJS{BJS}}
\def\ERS{Economic Research Service (ERS)
    \gdef\ERS{ERS}}
\def\USDA{U.S. Department of Agriculture (USDA)
    \gdef\USDA{USDA}}
\def\BEA{Bureau of Economic Analysis (BEA)
    \gdef\BEA{BEA}}

\def\NSO{national statistical office (NSO)
    \gdef\NSO{NSO}\gdef\NSOS{NSOs}}
\def\NSOS{national statistical offices (NSOs)
    \gdef\NSO{NSO}\gdef\NSOS{NSOs}}

\def\PI{Principal Investigator (PI)
    \def\PI{PI}}
\def\PIs{Principal Investigators (PIs)
    \def\PIs{PIs}}
\def\LEHD{Longitudinal Employer-Household Dynamics (LEHD)
    \def\LEHD{LEHD}}
\def\INSEE{Institut national de la statistique et des \'etudes \'economiques (INSEE)
    \def\INSEE{INSEE}}
\def\CREST{Centre de Recherche en \'Economie et Statistique (CREST)
    \def\CREST{CREST}}
\def\IAB{Institut f\"ur Arbeitsmarkt- und Berufsforschung (IAB)
     \def\IAB{IAB}}
\def\CBS{Centraal Bureau voor de Statistiek (CBS)
      \def\CBS{CBS}}
\def\MCU{multipoint conferencing units (MCU)
      \def\MCU{MCU}}

\def\CESCB{Center for Economic Studies (CES)
    \gdef\CESCB{CES}}

\def\SACNAS{Society for Advancement of Chicanos and Native Americans in Science (SACNAS)
    \gdef\SACNAS{SACNAS}}
\def\AWM{Association for Women in Mathematics (AWM)
    \gdef\AWM{AWM}}

\def\USPS{United States Postal Service (USPS)
    \gdef\USPS{USPS}}

\def\HSB{High School and Beyond (HS\&B:82)
    \gdef\HSB{HS\&B:82}}
\def\NELS{National Educational Longitudinal Study (NELS:88)
    \gdef\NELS{NELS:88}}
\def\NLS{National Longitudinal Study (NLS:72)
    \gdef\NLS{NLS:72}}
\def\NDI{National Death Index (NDI)
    \gdef\NDI{NDI}}

\def\NAS{National Academy of Sciences (NAS)
    \gdef\NAS{NAS}}
\def\EPRI{Electric Power Research Institute (EPRI)
    \gdef\EPRI{EPRI}}
\def\PCORI{Patient-Centered Outcomes Research Institute (PCORI)
    \gdef\PCORI{PCORI}}
\def\SHRP2{Strategic Highway Research Program 2 (SHRP 2)
    \def\SHRP2{SHRP 2}}
\def\OMOP{Observational Medical Outcomes Partnership (OMOP)
    \gdef\OMOP{OMOP}}
\def\ASM{Annual Survey of Manufactures (ASM)
    \gdef\ASM{ASM}}
\def\SIPP{Survey of Income and Program Participation (SIPP)
    \gdef\SIPP{SIPP}}

\def\NSRCG{National Survey of Recent College Graduates (NSRCG)
	\gdef\NSRCG{NSCRG}}
\def\SEH{science, engineering and health (SEH)
    \gdef\SEH{SEH}}
\def\HRS{University of Michigan Health and Retirement Study (HRS)
    \gdef\HRS{HRS}}
\def\NPSAS{National Postsecondary Student Aid Study (NPSAS)
    \gdef\NPSAS{NPSAS}}
\def\BnB{Baccalaureate and Beyond (B\&B)
    \gdef\BnB{B\&B}}
\def\BPS{Beginning Postsecondary Students Longitudinal Study (BPS)
    \gdef\BPS{BPS}}

\def\CMSS{Computation Methods in the Social Sciences (CMSS)
    \gdef\CMSS{CMSS}}

\def\TCS{Teacher Compensation Survey (TCS)
	\gdef\TCS{TCS}}
\def\SASS{Schools and Staffing Survey (SASS)
	\gdef\SASS{SASS}}
\def\NEA{National Education Association (NEA)
	\gdef\NEA{NEA}}
\def\FARS{Fatality Analysis and Reporting System (FARS)
	\gdef\FARS{FARS}}
\def\EDA{exploratory data analysis (EDA)
	\gdef\EDA{EDA}}

\def\CAT{computerized adaptive testing (CAT)
    \gdef\CAT{CAT}}
\def\CBT{computer-based testing (CBT)
    \gdef\CBT{CBT}}

\def\IRT{item response theory (IRT)
    \gdef\IRT{IRT}}

\def\CAPI{computer-assisted personal interview (CAPI)
    \gdef\CAPI{CAPI}}
\def\CATI{computer-assisted telephone interview (CATI)
    \gdef\CATI{CATI}}

\def\CAR{conditional autoregressive (CAR)
    \gdef\CAR{CAR}}

\def\IPUMS{Integrated Public Use Microdata Series (IPUMS)
    \gdef\IPUMS{IPUMS}}

\def\LEA{local education authority (LEA)
    \gdef\LEA{LEA}\gdef\LEAS{LEAs}}
\def\SEA{state education authority (SEA)
    \gdef\SEA{SEA}\gdef\SEAS{SEAs}}
\def\LEAS{local education authorities (LEAs)
    \gdef\LEA{LEA}\gdef\LEAS{LEAs}}
\def\SEAS{state education authorities (SEAs)
    \gdef\SEA{SEA}\gdef\SEAS{SEAs}}

\def\CMS{Centers for Medicare and Medicaid Services (CMS)
    \gdef\CMS{CMS}}

\def\MMS{Methodology, Measurement, and Statistics (MMS)
    \gdef\MMS{MMS}}

\def\SOC{standard occupational code (SOC)
    \gdef\SOC{SOC}}
\def\ASCO{American Society for Clinical Oncology (ASCO)
    \gdef\ASCO{ASCO}}

\def\UI{unemployment insurance (UI)
    \gdef\UI{UI}}

\def\PT{Project TALENT (PT)
    \gdef\PT{PT}}

\def\AERA{American Educational Research Association (AERA)
    \gdef\AERA{AERA}}

\def\CBSA{Core Business Statistical Area (CBSA)
    \gdef\CBSA{CBSA}\def\CBSAs{CBSAs}}
\def\CBSAs{Core Business Statistical Areas (CBSAs)
    \gdef\CBSA{CBSA}\def\CBSAs{CBSAs}}

\def\PT{Project Talent (PT)
    \gdef\PT{PT}}
\def\FCB{First Citizens Bank (FCB)
    \gdef\FCB{FCB}}
\def\BWF{Burroughs Wellcome Fund (BWF)
    \gdef\BWF{BWF}}

\def\PPRL{privacy-preserving record linkage (PPRL)
    \gdef\PPRL{PPRL}}

\def\NIA{National Institute on Aging (NIA)
    \gdef\NIA{NIA}}

\def\CHAID{Chi-squared automatic interaction detection (CHAID)
    \gdef\CHAID{CHAID}}
\def\GSC{General School Characteristics (GSC)
    \gdef\GSC{GSC}}

\def\CoDA{Center of Excellence for Complex Data Analysis (CoDA)
    \gdef\CoDA{CoDA}}

\def\AIC{Akaike information criterion (AIC)
    \gdef\AIC{AIC}}
\def\BIC{Bayes information criterion (BIC)
    \gdef\BIC{BIC}}
\def\SSES{Social, Statistical, and Environmental Sciences (SSES)
    \gdef\SSES{SSES}}
\def\SCSS{Survey, Computing, and Statistical Sciences (SCSS)
    \gdef\SCSS{SCSS}}
\def\DSDS{Division for Statistical and Data Science (DSDS)
    \gdef\DSDS{DSDS}}

\def\CM{Census of Manufactures (CM)
    \gdef\CM{CM}}

\def\FSRDC{Federal Statistical Research Data Center (FSRDC)
    \gdef\FSRDC{FSRDC}\gdef\FSRDCS{FSRDCs}}
\def\FSRDCS{Federal Statistical Research Data Centers (FSRDCs)
    \gdef\FSRDC{FSRDC}\gdef\FSRDCS{FSRDCs}}

\def\PTTP{partially trusted third party (PTTP)
    \gdef\PTTP{PTTP}\gdef\PTTPS{PTTPs}}
\def\PTTPS{partially trusted third parties (PTTPs)
    \gdef\PTTP{PTTP}\gdef\PTTPS{PTTPs}}

\def\NAICS{North American Industrial Classification System (NAICS)
    \gdef\NAICS{NAICS}}
\def\CART{classification and regression trees (CART)
    \gdef\CART{CART}}

\def\FEBRL{Freely Extensible Biomedical Record Linkage (FEBRL)
    \gdef\FEBRL{FEBRL}}
\def\FRIL{Fine-Grained Records Integration and Linkage Tool (FRIL)
    \gdef\FRIL{FRIL}}
\def\SOEMPI{Secure Open Enterprise Master Patient Index (SOEMPI)
    \gdef\SOEMPI{SOEMPI}}

\def\SDA{statistical disclosure avoidance (SDA)
    \gdef\SDA{SDA}}

\def\LAS{Laboratory for Analytic Sciences (LAS)
    \gdef\LAS{LAS}}

\def\IPF{iterative proportional fitting (IPF)
    \gdef\IPF{IPF}}
\def\ROC{receiver operating characteristic (ROC)
    \gdef\ROC{ROC}}

\def\NCBI{National Center for Biotechnology Information (NCBI)
    \gdef\NCBI{NCBI}}

\def\SCC{Specified Certainty Classification (SSC)
    \gdef\SCC{SCC}}

\def\BP{base pairs (BP) 
    \gdef\BP{BP}}

\def\ROC{receiver operating characteristic (ROC)
    \gdef\ROC{ROC}}

\gdef\MV{\texttt{Mason\_variator}}

\def\AI{artificial intelligence (AI)
   \gdef\AI{AI}}

\def\SARS{severe acute respiratory syndrome (SARS)
  \gdef\SARS{SARS}}

\def\SNP{single nucleotide polymorphism (SNP)
  \gdef\SNP{SNP}
  \gdef\SNPs{SNPs}}

\def\SNPs{single nucleotide polymorphisms (SNPs)
  \gdef\SNPs{SNPs}
  \gdef\SNP{SNP}}

\def\CRISPR{clustered regularly interspaced short palindromic repeats (CRISPR)
  \gdef\CRISPR{CRISPR}}

\def\CAS{CRISPR associated sequence (CAS)
  \gdef\CAS{CAS}}

\def\MDS{multidimensional scaling (MDS)
  \gdef\MDS{MDS}}

\begin{document}
\title{Measuring Quality of DNA Sequence Data via Degradation}

\author{Alan F. Karr\thanks{Fraunhofer USA CMA; akarr@fraunhofer.org; Corresponding author}, Jason Hauzel\thanks{Fraunhofer USA CMA; jhauzel@fraunhofer.org}, Adam A. Porter\thanks{University of Maryland, Department of Computer Science; aporter@cs.umd.edu}, Marcel Schaefer\thanks{Fraunhofer USA CMA; mschaefer@fraunhofer.org}}

\maketitle

\begin{abstract}
We propose and apply a novel paradigm for characterization of genome data quality, which quantifies the effects of intentional degradation of quality. The rationale is that the higher the initial quality, the more fragile the genome and the greater the effects of degradation. We demonstrate that this phenomenon is ubiquitous, and that quantified measures of degradation can be used for multiple purposes. We focus on identifying outliers that may be problematic with respect to data quality, but might also be true anomalies or even attempts to subvert the database.

\end{abstract}

\section{Introduction}\label{sec.intro}
As public genome databases proliferate, their immense scientific power is tempered by skepticism about their quality. The skepticism is not merely anecdotal: there are documented instances and implications \citep{critical-assessment-2021, langdon-2014, steinegger-2020}. Although we argue in Appendix \ref{app.dq} that data quality should not be construed as comprising only errors in data, the principal contribution of the paper is a novel paradigm for measuring quality of genome sequences by deliberately introducing errors that reduce quality, a process we term degradation. The errors are \SNPs, insertions and deletions that both occur naturally as mutations and arise in next generation sequencing. Our reasoning is that \emph{higher quality data are more fragile}: the higher the initial quality, the greater the effect of the same amount of degradation. We present evidence that supports this reasoning, as well as demonstrates the scope and consequences of the phenomenon.

Even though the main contribution of the paper is methodological, applicability to bioinformatics problems is its \emph{raison d'\^{e}tre}. Our exemplar problem is detection of outliers in genome databases: we identify genomes in a 26,953 coronavirus database whose degradation behavior is anomalous, and whose quality, therefore, may be suspect. We detect deliberately inserted low quality genomes, but other genomes in the original database are equally problematic. A second potential application is to thwart adversarial attacks on genome databases that, for instance, insert artificial genomes so that sequences of concern such as those generated by the methods in \cite{farbiash-puzis-2020} will pass screening tests. Finally, degradation can be used to characterize the quality of synthetic DNA reads that are used to evaluate genome assemblers \citep{hybrid-reads-2020}.

Our method is rooted in total quality paradigms for official statistics, that is, censuses and surveys conducted by national statistics offices (see Appendix \ref{app.dq}). In that context, data quality is a longstanding issue, and low quality data are known to be resistant to further errors, such as those introduced by editing, imputation or \SDL. We also draw on official statistics for techniques to quantify data quality. In experimental settings and because it is intuitive, we measure degradation by distance, appropriately defined, from the stating point. In real databases, this is not possible, so we employ measures of distance from a universal ``endpoint'' representing the lowest possible quality---pure randomness in the form of maximal entropy, which every genome reaches in the limit of infinite degradation.

The remainder of the paper is organized as follows. To maintain flow, brief background on data quality appears in Appendix \ref{app.dq}. Our hypothesis that higher quality means greater fragility and initial evidence supporting it appear in \S\ref{sec.argumentation}. In \S\ref{sec.bases}, we examine degradation of tuple distributions, linking this paper to work on applying the Markov structure of genome sequences \citep{markovstructure-2021}. A central degradation measure---entropy of triplet distributions---is applied in \S\ref{sec.outliers} to outlier detection. In \S\ref{sec.extensions}, we demonstrate the effects of degradation on higher-level genome structure, \emph{viz}, repeats and palindromes, and on non-virus genomes. Conclusions appear in \S\ref{sec.conclusions}.

\section{Argumentation}\label{sec.argumentation}
Here are our central hypothesis and initial evidence supporting it.

\subsection{The Hypothesis}\label{subsec.hypothesis}
We hypothesize that because high quality data are more fragile than low quality data,\footnote{As noted, there is precedent in official statistics for this assertion. Some components of the argument appear in \cite{dq-statmeth06}, while the \TSE\ paradigm referred to in Appendix \ref{app.dq} rests in part on this premise.} the quality of elements of a DNA sequence database can be measured by degrading them, Secondarily, the more complex the characteristic examined, the greater the impact of degradation. As we see below, the effect of degradation increases as we move from base distributions to pair distributions to triplet distributions to quartet distributions to repeats and palindromes.

We perform the degradation by iteratively applying the \MV\ software \citep{fu_mi_publications962}. Briefly, the \MV\ simulates changes to a genome sequence: \SNPs, insertions, deletions, inversions, translocations, and duplications, with specified probabilities for each. Such changes occur naturally as mutations as well as in reads produced by next generation sequencers, such as those made by Illumina. \MV\ runs from a command line interface with user-specified parameters, input files, and output files. For simplicity, in most of our analyses, only \SNPs\ were simulated.\footnote{The principal reason is to avoid burdensome computation of Levenshtein distances.} Iterative use of \MV\ means starting with a given genome, running \MV\ on it, running \MV\ again on the result, \dots, up to a specified number of iterations, which is usually 2000.\footnote{Much the same effect could be achieved by increasing the error probabilities, but at a loss of interpretability, because parametrization by the number of iterations is more intuitive.}

We need measures of quality for degraded genomes, and have investigated several. The first two of these are employed in string matching: Hamming distance \citep{navarro-string-matching-2001} is usable when only \SNPs\ are simulated, while Levenshtein distance \citep{navarro-string-matching-2001}  allows insertions and deletions that alter the length of the DNA sequence.\footnote{The Hamming distance between sequences with different lengths is infinite. Levenshtein distance is significantly more burdensome than Hamming distance computationally, with respect to both time and memory requirements, especially for longer genomes.} As discussed below, the origin point for Hamming and Levenshtein distances is crucial. Distances based on distributions of nucleotides, pairs, triplets and quartets are discussed in \S\ref{sec.bases}. Entropy of triplet distributions of degraded genomes is examined in detail in \S\ref{subsec.bd-degradation} and forms the basis of \S\ref{sec.outliers}.

\subsection{Initial Evidence}\label{subsec.initialevidence}
\begin{figure}[htbp]
\begin{center}
\includegraphics[width=4in]{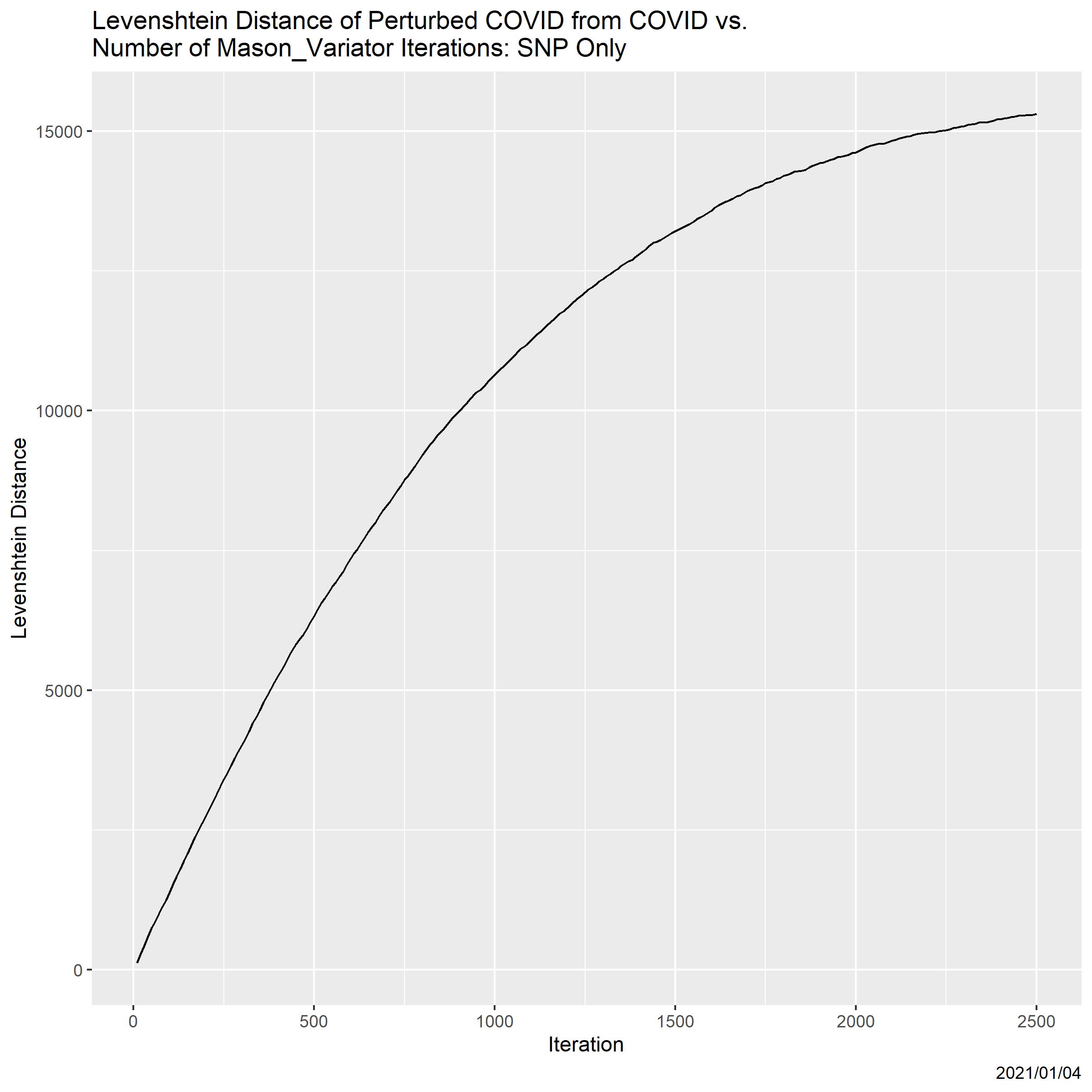}
\end{center}
\caption{Levenshtein distance from a (randomly selected) coronavirus genome as a function of the number of \MV\ iterations.}
\label{fig.concept}
\end{figure}

Figure \ref{fig.concept} visualizes the hypothesis for a single element of the coronavirus database employed in \S\ref{sec.bases} and \ref{sec.outliers}. In the figure, the $x$-axis is the number of iterations of the \MV, and the $y$-axis is Levenshtein distance between the degraded genome and the original genome. All forms of errors were allowed. The most salient characteristic of the curve is its concavity: the more degradation already done, the less the effect of each additional iteration.

\begin{figure}[htbp]
\begin{center}
\includegraphics[width=2.7in]{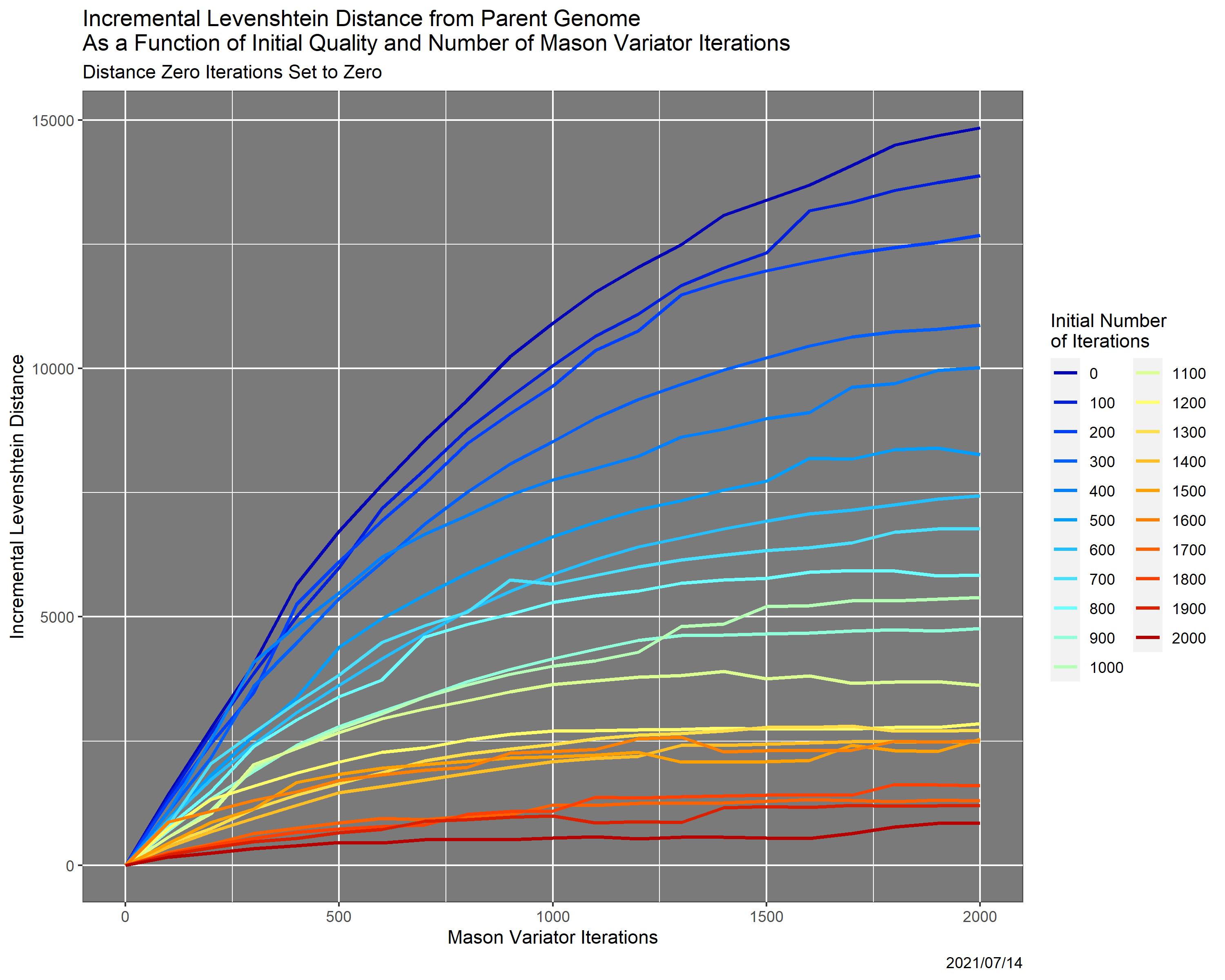}

\vspace{.1in}
\includegraphics[width=2.7in]{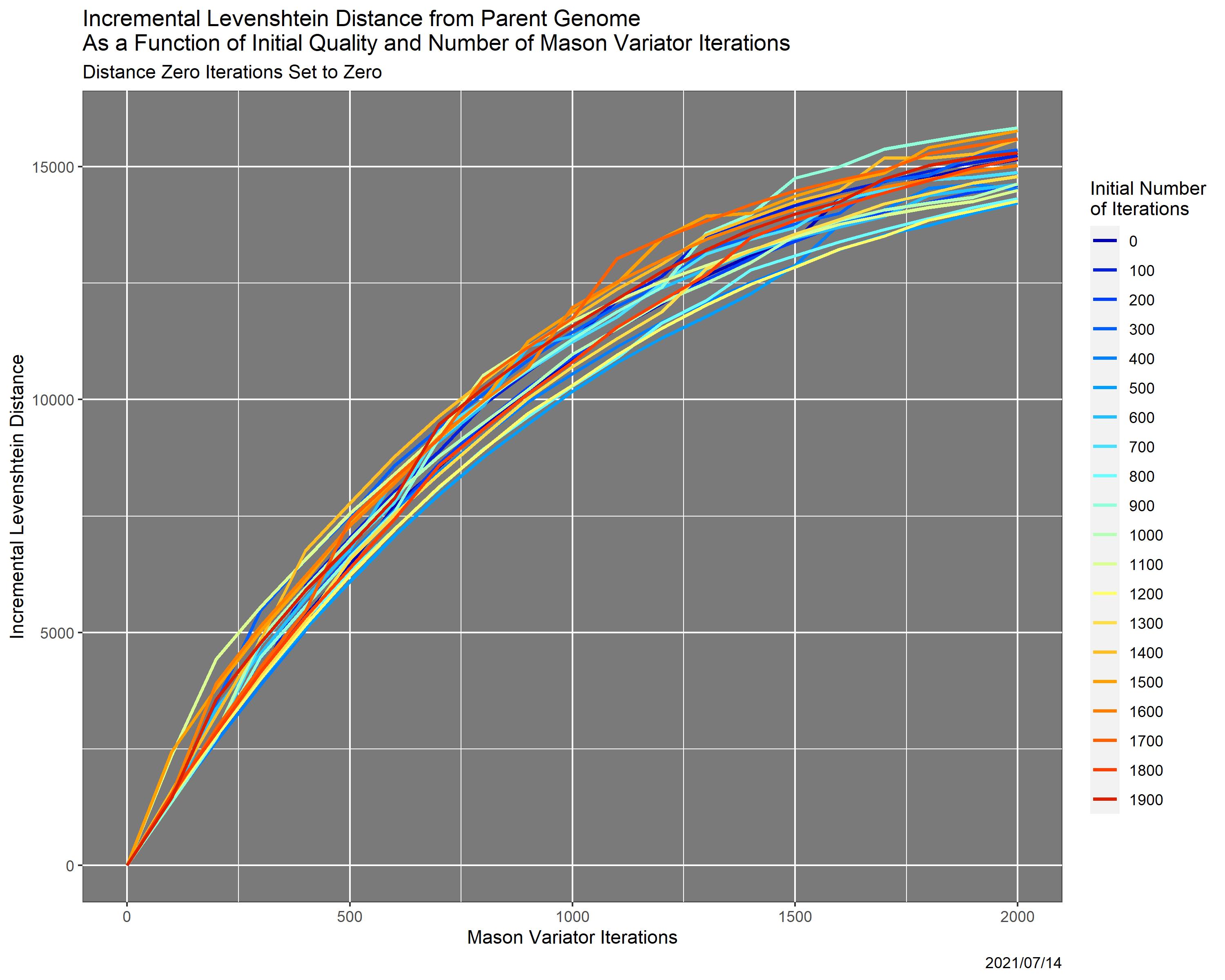}

\vspace{.1in}
\includegraphics[width=2.7in]{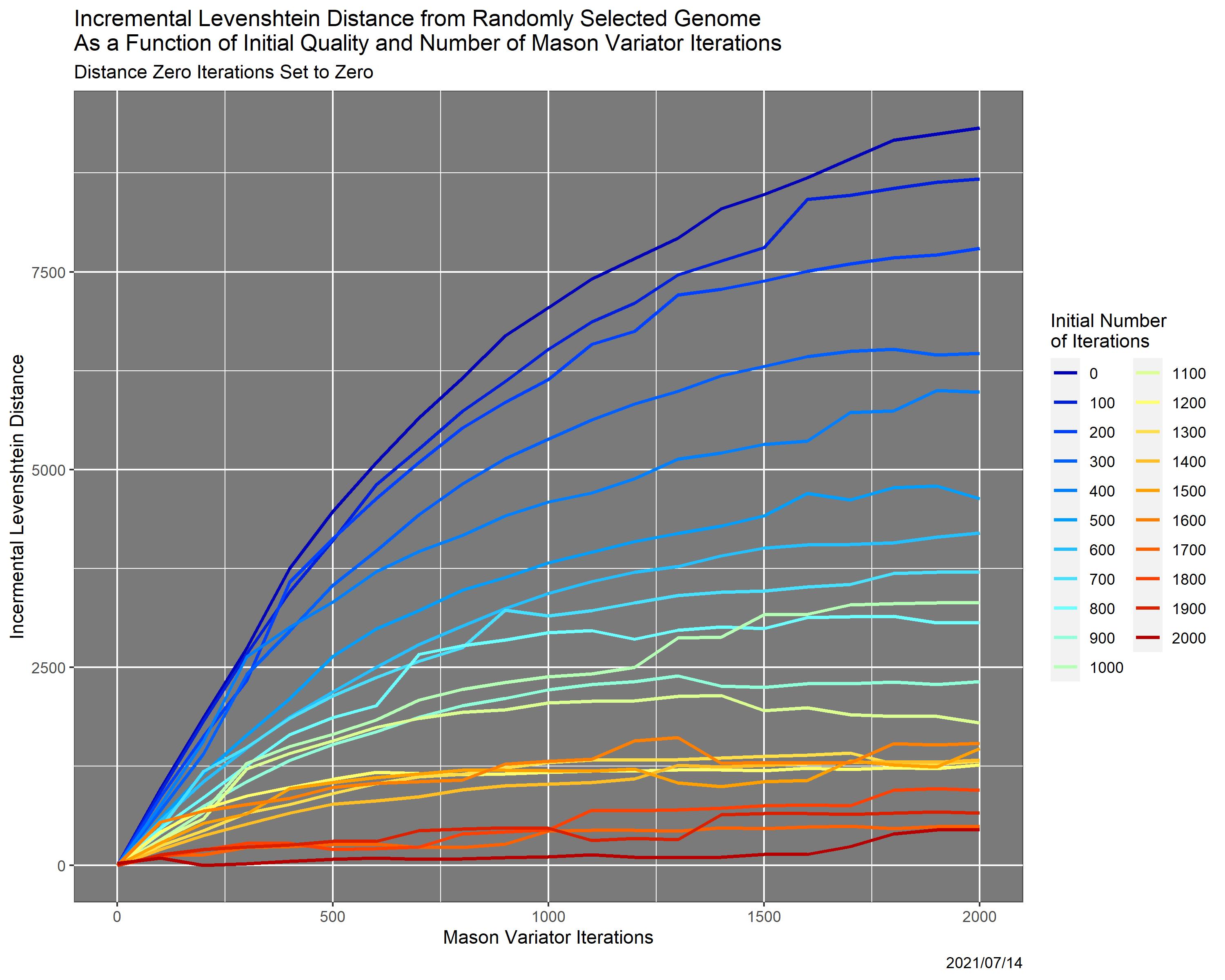}
\end{center}
\caption{Degradation behavior as a function of \MV\ iterations and initial data quality, for a randomly selected coronavirus genome. \emph{Top:} Degradation is measured by incremental Levenshtein distance from the parent (0-iteration) genome. \emph{Middle:} Degradation is measured by Levenshtein distance from the starting point. \emph{Bottom:} Degradation is measured by incremental Levenshtein distance from a second randomly selected genome in the \NCBI\ database. Color encodes the initial number of \MV\ iterations.}
\label{fig.starts}
\end{figure}

There are issues with the choice of the origin for Levenshtein distances. In Figure \ref{fig.starts}, there are 21 initial genomes---the one randomly selected coronavirus genome and that genome after $100, 200, \dots, 2000$ \MV\ iterations, representing continually decreasing initial data quality. In the top panel, Levenshtein distance is measured from the parent (0-iteration) genome, and the distance at iteration 0 has been subtracted from each curve. In a sense, however, this is ``cheating,'' because in real databases, there are not known parent genomes. In the middle panel in Figure \ref{fig.starts}, Levenshtein distance for each curve is measured from its starting point. The curves there differ little, and certainly not systematically. Fortunately, a work-around exists: the bottom panel in Figure \ref{fig.starts} shows that (within reason), any fixed genome can be used as the origin. There, all Levenshtein distances are measured from a second randomly selected genome in the \NCBI\ dataset. The key point is that the curves in the bottom panel vary significantly and systematically with respect to the initial degradation.

In \S \ref{subsec.bd-triplets}, we instead measure distance from a fixed ``endpoint'' interpretable as infinite degradation.


\section{Degradation of Tuple Distributions}\label{sec.bases}
Distributions of bases, pairs of successive bases, triplets of successive bases and quartets of successive bases differ across genomes, in ways that support a variety of analyses, including not only outlier identification (\S\ref{subsec.outliers-bd-clustering}), but also Bayesian classification of simulated Illumina reads and detection of contamination \citep{markovstructure-2021}. How tuple distributions behave under degradation supports our hypothesis that application of the \MV\ decreases data quality. Higher-level genome structure such as repeats and palindromes is discussed in \S\ref{subsec.dnastructure}.

As noted in \S\ref{sec.intro}, our experimental platform is a coronavirus database containing 26,953 genomes, which was downloaded from the \NCBI\ in November of 2020. To it, we added eleven ''known'' problem cases: a single adenovirus genome with length 34,125 BP (downloaded as part of the \texttt{Art} read simulator software package \citep{art-2012}) and low-quality versions of 10 coronavirus selected randomly from the original 26,953, each created by 2000 iterations of the \MV. Our method detects not only these known outliers---the minimal criterion for credibility, but also others.

\subsection{Preliminaries}\label{subsec.bd-preliminaries}
In this paper, a genome $G$ is a character string chosen from the alphabet $\mathcal{B} = \{A, C, G, T\}$, and represents one strand of the DNA\footnote{Or, for viruses, RNA.} in an organism. The constituent bases (nucleotides) are A = adenine, C = cytosine, G = guanine and T = thymine. We denote the length of a genome $G$ by $|G|$; the $i^{\mathrm{th}}$ base in $G$ is $G(i)$; and the bases from location $i$ to location $j > i$ are $G(i:j)$. Given an integer $n \geq 1$, the $n$-gram distribution is the probability distribution $P_n(\cdot|G)$ on the set all sequences of length $n$ chosen from $\mathcal{B}$---there are $4^n$ of them---constructed by forming a table of all length $n$ contiguous substrings of $G$ and normalizing it so that its entries sum to 1.\footnote{There are $|G|-n+1$ such sequences, starting at $1, 2, \dots, |G|-n+1$, so the normalization amounts to division by $|G|-n+1$.} In \S\ref{sec.outliers}, we focus on \emph{triplets}, which are 64-dimensional summaries of genomes, and which also encode amino acids---the building blocks of proteins. The interpretation of $P_3(\cdot|G)$ is that for each choice of $b_1, b_2, b_3$ from $\mathcal{B}$,
\begin{equation}
P_3(b_1b_2b_3|G) = \mathrm{Prob}\{G(k:[k+2]) = b_1b_2b_3\},
\label{eq.tripletdistribution}
\end{equation}
where $k$ is chosen at random from $\{1, \dots, |G|-2\}$. Triplets provide a generative model of a genome as a second-order Markov process, since $P_3$ contains the same information as the pair distribution $P_2$ and the $16 \times 4$ transition matrix
\begin{equation}
T_3(b_1, b_2, b_3|G) = \mathrm{Prob}(G(k+2) = b_3| G(k) = b_1, G(k+1) = b_2)
\label{eq.transitionmatrix3}
\end{equation}
that gives the distribution of each base conditional on its two immediate predecessors.

Other cases of interest are less suited to our purposes. Base ($n=1$) and pair ($n=2$) distributions are too coarse to be useful on their own. Quartets ($n=4$) have been studied extensively \citep{pride-tetra-2003, teeling-tetra-classification-2004, teeling-tetra-web-2004}. For the problems we address, they are more cumbersome than triplets without being significantly more informative. Finally, although we do not do so here, triplet distributions can usefully be converted to amino acid distributions \citep{markovstructure-2021}.

\subsection{General Effects of Degradation}\label{subsec.bd-degradation}
\begin{figure}[htbp]
\begin{center}
\includegraphics[width=4in]{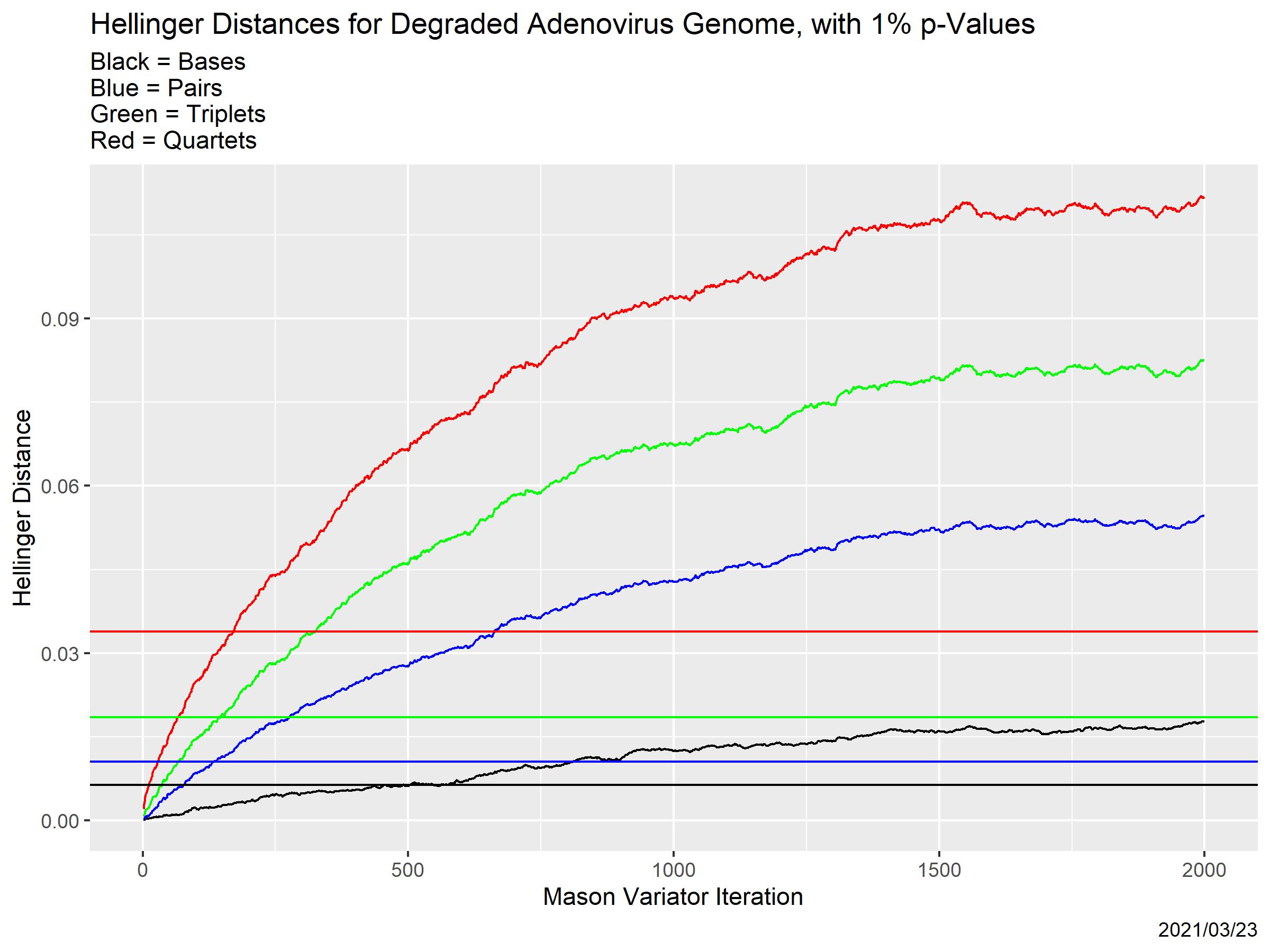}
\end{center}
\caption{Effect of degradation on base distribution (black), pair distribution (blue), triplet distribution (green) and quartet distribution (red) in the adenovirus genome.}
\label{fig.degradedbasedistributions}
\end{figure}

For the adenovirus genome, Figure \ref{fig.degradedbasedistributions} shows the effect of degradation on the base, pair, triplet and quartet distributions, measured by Hellinger distance \citep{nikulin-hd-2010} from corresponding distributions for the original genome. The interpretation is that as the number of \MV\ iterations increases, base, pair, triplet distributions, and quartet distributions all move farther and farther away from the parent genome, at slower and slower rates. Moreover, confirming the secondary hypothesis in \S\ref{subsec.hypothesis}, the higher the dimensionality, the more rapid the movement: quartets are more fragile than triplets, which are more fragile than pairs, which are more fragile than individual bases. The horizontal lines in Figure \ref{fig.degradedbasedistributions}, whose colors match those of the curves, are simulation-derived 1\% $p$-values: the probability that the distribution matches that of the original genome is less than 0.01 when the distance exceeds the line. Interestingly, the numbers of iterations at which the 1\% thresholds are passed (i.e., where the curves cross the lines) are nearly the same for pairs, triplets and quartets and lower than for bases alone.

\subsection{Measuring Degradation of Triplet Distributions via Entropy}\label{subsec.bd-triplets}
So far, we have confined attention to what degradation moves away from. In many ways, what it moves \emph{toward} is more useful, because there is a single infinite degradation endpoint representing maximum entropy, which in statistical physics and probability theory, measures disorder. The entropy of a probability distribution $P$ on a finite set $\mathcal{S}$ is
\begin{equation}
H(P) = - \sum_{s \in \mathcal{S}} p(s) \log p(s),
\label{eq.entropy}
\end{equation}
with the convention that $0 \times -\infty = 0$. Entropy is minimized by distributions concentrated at a single point and maximized at the uniform distribution on $\mathcal{S}$, with maximizing value $\log(|\mathcal{S}|)$, where $|\mathcal{\cdot}|$ denotes cardinality. The existence of the universal maximizing value enables us to measure degradation as \emph{movement toward maximum entropy}, removing the common origin issue discussed in \S\ref{subsec.initialevidence}.

\begin{figure}[htbp]
\begin{center}
\includegraphics[width=4in]{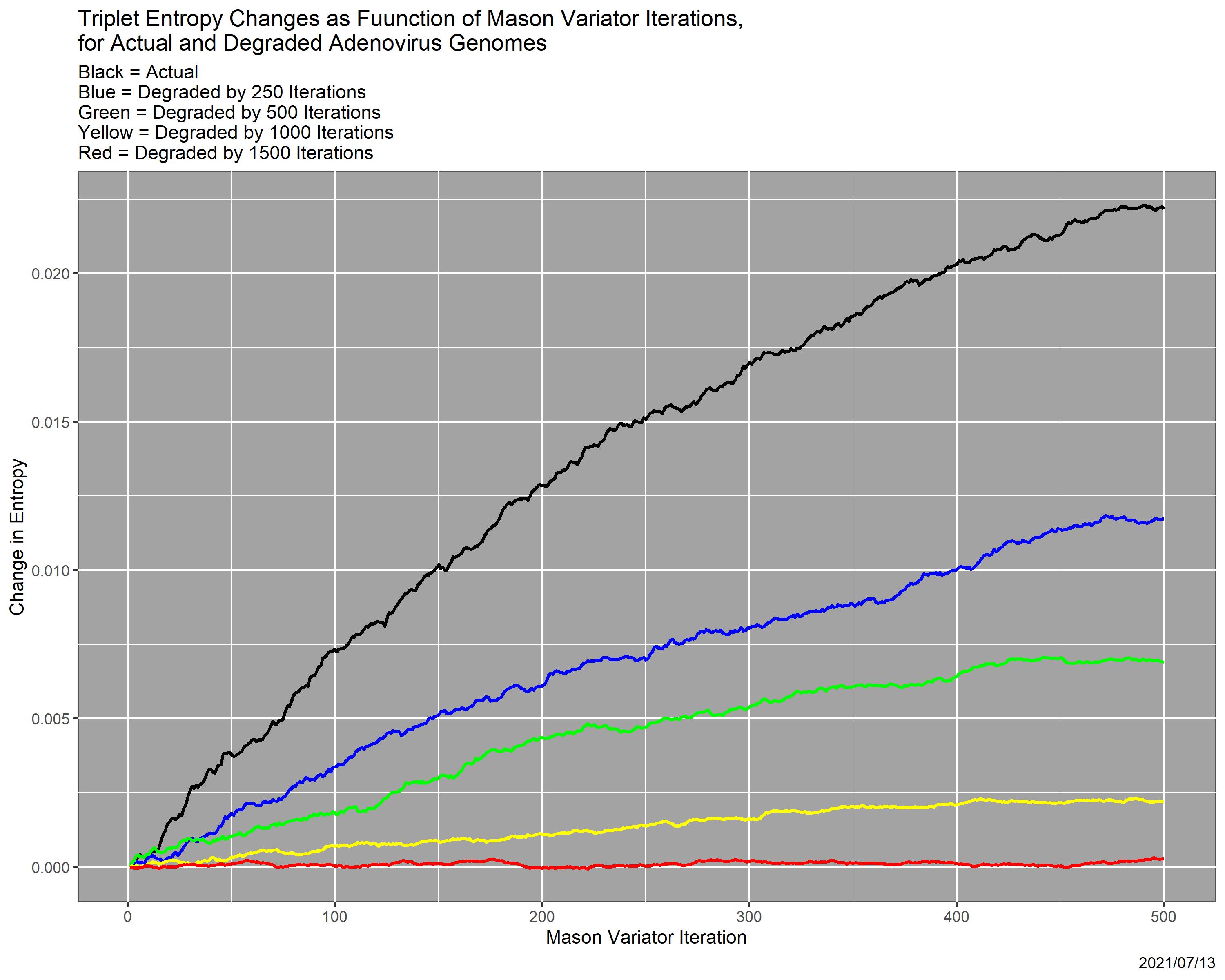}
\end{center}
\caption{Change in entropy as a function of \MV\ iterations, for one adenovirus genome and four degraded versions of it. \emph{Black:} original genome; \emph{Blue:} degraded by 250 iterations; \emph{Green:} degraded by 500 iterations; \emph{Yellow:} degraded by 1000 iterations; \emph{Red:} degraded by 1500 iterations.}
\label{fig.startingpoints2}
\end{figure}

Figure \ref{fig.startingpoints2} shows the effect of 500 \MV\ iterations on entropy of triplet distributions---hereafter, just triplet entropy---of the adenovirus genome, starting from the genome itself (black curve) compared to starting from the genome degraded by 250 \MV\ iterations (blue curve), degraded by 500 MV iterations (green curve), degraded by 1000 \MV\ iterations (yellow curve), and degraded by 1500 \MV\ iterations (red curve). The $y$-axis is the increase in entropy as a function of \MV\ iterations, so Figure \ref{fig.startingpoints2} shows movement toward maximal entropy.

\begin{figure}[htbp]
\begin{center}
\includegraphics[width=4in]{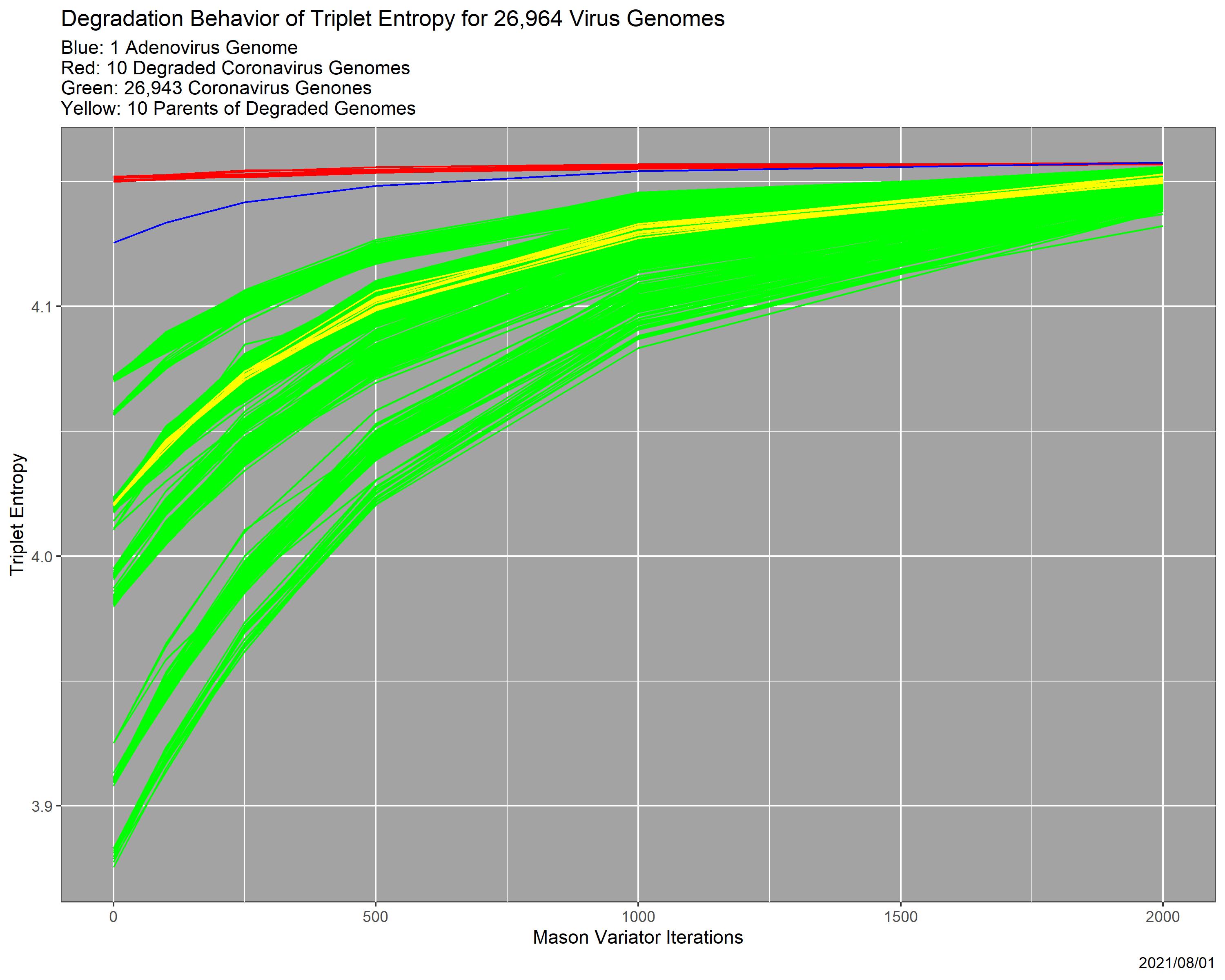}
\end{center}
\caption{Entropy as a function of \MV\ iterations, for the 26,964-genome database. The adenovirus outlier is in blue, the 10 degraded coronavirus outliers are in red, and the parents of the ten coronavirus outliers are in yellow.}
\label{fig.entropydegradation-26964}
\end{figure}

Figure \ref{fig.entropydegradation-26964} shows, albeit with massive overplotting, the triplet entropy degradation for the entire 26,964-element experimental database. The adenovirus genome, in blue, and the 10 degraded coronavirus genomes, in red, are apparent outliers. But, clearly there are also other outliers, which we pursue in \S\ref{subsec.outliers-deg-clustering}.

\section{Outlier Detection}\label{sec.outliers}
One effective strategy for identifying elements of a genome database with problematic quality is to search for outliers. We concede that the implicit presumption that the bulk of the database is of high quality may be untested. The key question is, ``Outlying with respect to what metric?'' In this section, the metric is based on hierarchical cluster analysis of triplet entropy increase resulting from \MV\ degradation, that is, on the shapes of the curves in Figure \ref{fig.entropydegradation-26964}. The clustering is in three dimensions, as opposed to 64 dimensions for triplet distributions and 21 for amino acid distributions. Possibly unexpectedly, the results in \cite{markovstructure-2021} and those in \S\ref{subsec.outliers-deg-clustering} are very similar.

\subsection{Outlier Detection Using Triplet Distributions}\label{subsec.outliers-bd-clustering}
We showed in \cite{markovstructure-2021} that clustering of triplet distributions identifies outliers. Briefly, we performed
hierarchical clustering, using Euclidean distances and ``complete clustering'' in \texttt{R} \citep{R-2021}, on the 26,964-genome database, using as clustering variables the 64 standardized components of the triplet distributions. By means of standard heuristics that trade off model fit and model complexity, the number of clusters was determined to be 23.

\begin{figure}[htbp]
\begin{center}
\includegraphics[width=5in]{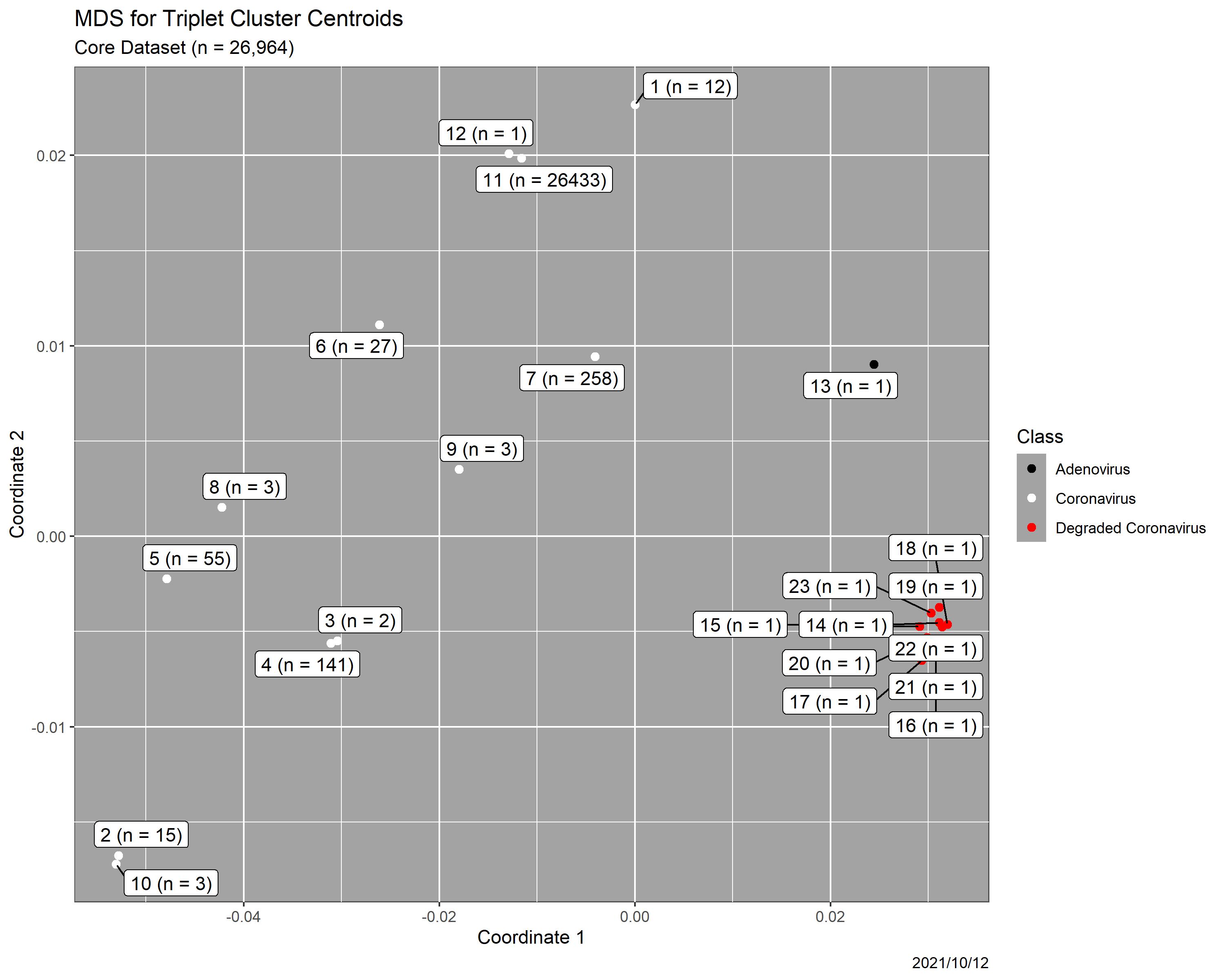}

\vspace{.1in}
\includegraphics[width=5in]{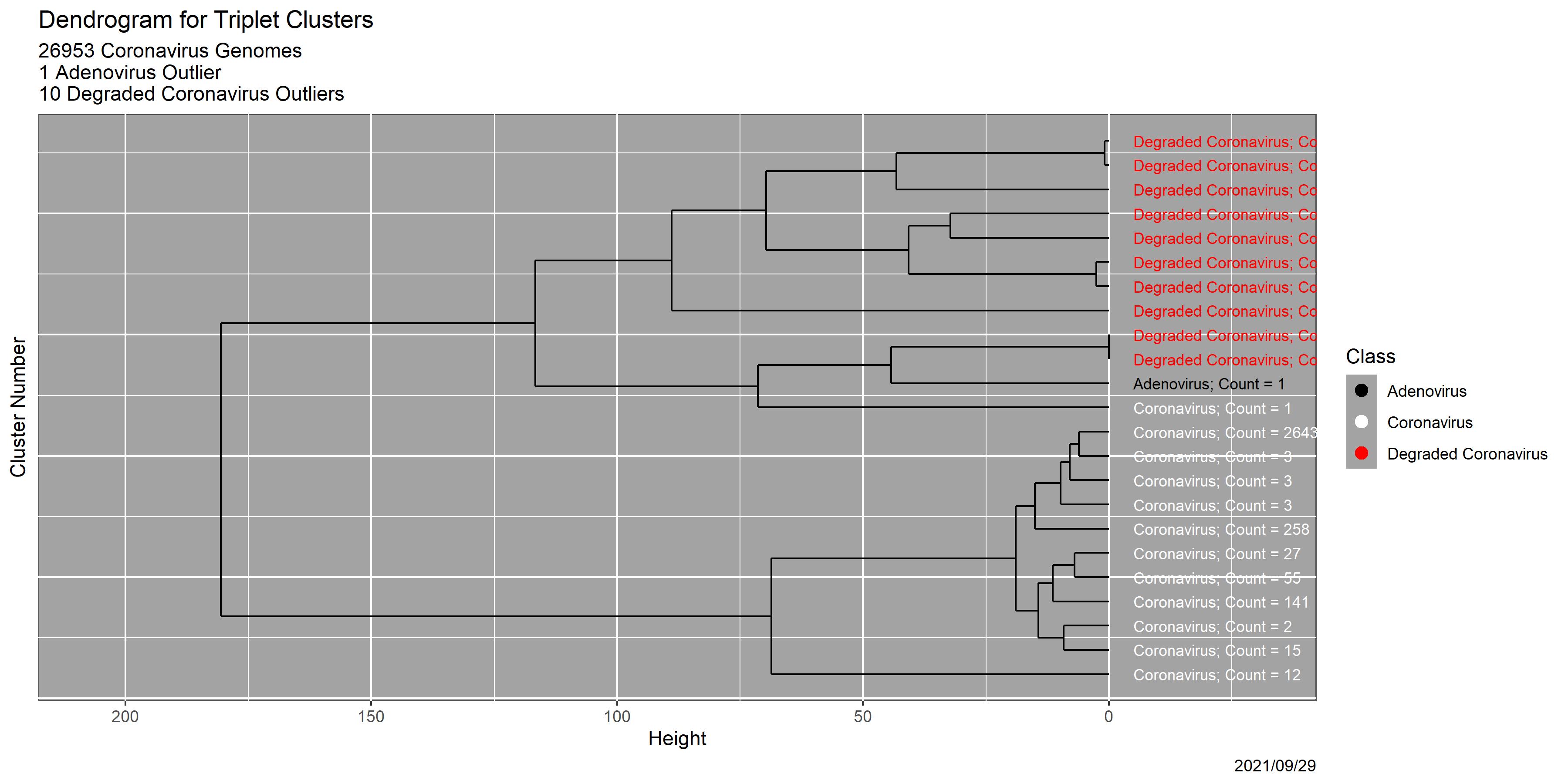}
\end{center}
\caption{\emph{For triplet distribution clustering, Top:} two-dimensional \MDS\ plot of the 23 triplet distribution cluster centroids for a 26,964-element coronavirus database containing 11 deliberately added outliers. Labels are cluster numbers and counts. Greater distance implies higher dissimilarity. \emph{Bottom:} Corresponding dendrogram, in which clusters are in order from 1 at the bottom to 23 at the top. Colors are the same in both graphics.}
\label{fig.triplets-clusters}
\end{figure}

The top panel in Figure \ref{fig.triplets-clusters} is a plot of two-dimensional \MDS\ \citep{kruskal-mds-1964, cox-cox-mds-2001} of the 23 cluster centroids. The overwhelming majority of coronavirus genomes---26,433 of the original 26,953, or 98.1\%---are in a single cluster. One original coronavirus genome appears by itself, in cluster 12. Cluster 13 contains the adenovirus genome alone, while each of the 10 degraded coronavirus genomes appears in a cluster by itself (clusters 14--23). Thus, the deliberate outliers are not only detected but also distinguished from one another. The dendrogram in the bottom panel of Figure \ref{fig.triplets-clusters} shows that the coronavirus genome in cluster 12 and the deliberate outliers are separated from the remaining 26,952 coronavirus genomes at the first split in the clustering process. Clusters 1--10, which are small, are potential outliers as well. See \cite{markovstructure-2021} for details and a scientific interpretation.

\subsection{Outlier Detection Using Entropy Degradation}\label{subsec.outliers-deg-clustering}
Here we cluster the genomes on the basis of degradation behavior of triplet entropy, that is, the curves shown in Figure \ref{fig.entropydegradation-26964}. As noted already, the results match closely with those in \S\ref{subsec.outliers-bd-clustering}.

The clustering is now in only three dimensions, reached by a path that starts with Figure \ref{fig.entropydegradation-26964}. Every one of the 26,964 curves plotted there is based on entropy following 0, 250, 500, 1000 and 2000 \MV\ iterations. We fitted a quadratic function to each set of 5 values, reducing the dimension to 3. These quadratic models are uniformly good: the smallest of the coefficients of determination, $R^2$, is 0.941 and 99\% of them exceed 0.976. Hierarchical clustering was then performed on standardized versions of the three quadratic coefficients, using the ``ward.D'' option in \texttt{R}, resulting in 34 clusters, with counts ranging from 11 to 1470. Statistically, the clustering is extremely good: the cluster numbers alone explain 98.96509\%, 99.02426\% and 98.45677\% of the variation in the quadratic coefficients.

\begin{figure}[htbp]
\begin{center}
\includegraphics[width=4in]{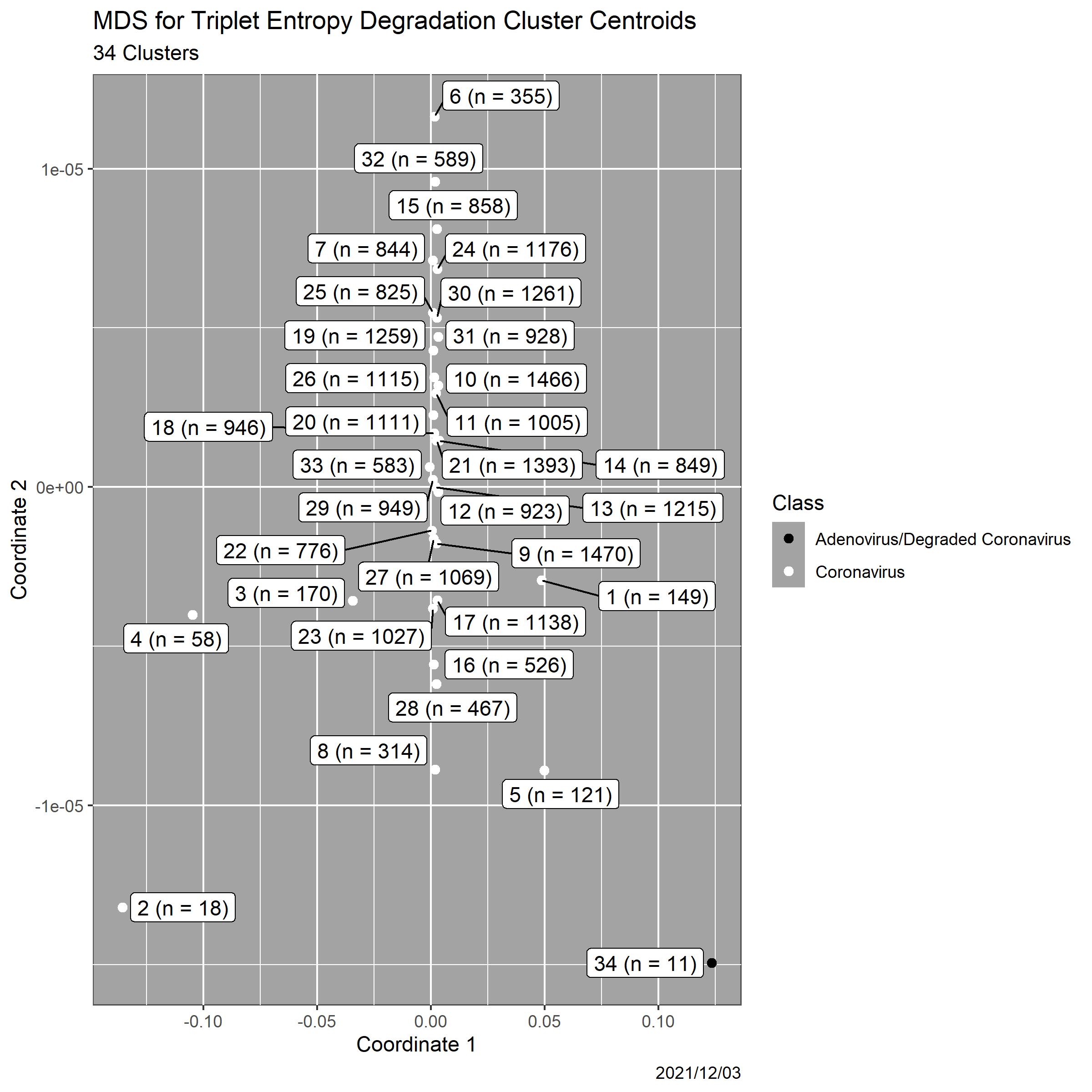}

\vspace{.1in}
\includegraphics[width=4.5in]{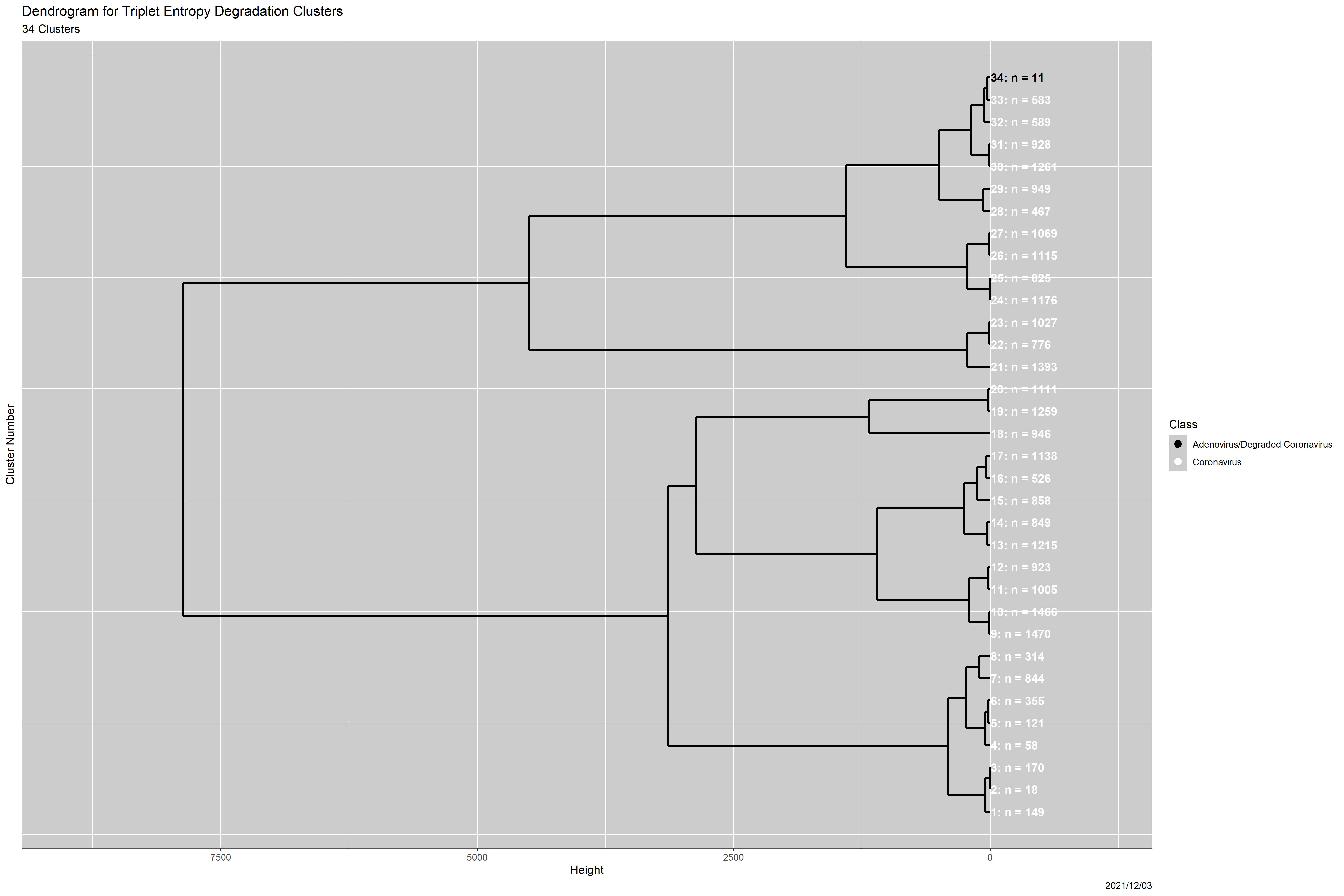}
\end{center}
\caption{\emph{For triplet entropy clustering, Top:} Two-dimensional \MDS\ plot of the 34 cluster centroids. Labels are cluster numbers and counts. Greater distance implies higher dissimilarity. \emph{Bottom:} Corresponding dendrogram, in which clusters are in order from 1 at the bottom to 34 at the top. Colors are the same in both graphics.}
\label{fig.entropy-clusters}
\end{figure}

Paralleling Figure \ref{fig.triplets-clusters}, Figure \ref{fig.entropy-clusters} shows the result of applying two-dimensional \MDS\ to the cluster centroids, as well as the associated dendrogram. There is nothing comparable to the massive coronavirus cluster in the triplet distribution analysis. As noted above, the largest cluster in the triplet entropy degradation cluster contains only 1470 genomes. The adenovirus outlier and the ten degraded coronavirus outliers are placed together in cluster 34, and Figure \ref{fig.entropy-clusters} shows that they clearly differ from all of the other genomes. Clusters 1--5 are candidate outliers. Not only are they relatively small, but also each differs strongly from \emph{all} of the other clusters. They are suggestively similar to clusters 1--10 in Figure \ref{fig.triplets-clusters}, which we pursue in \S\ref{subsec.outliers-relationships}. In Figure \ref{fig.entropy-clusters}, the 11 deliberate outliers in cluster 34 are distant from the majority of the coronavirus genomes, but no more so than the 18 coronavirus genomes in cluster 2.

\subsection{Relationships Between the Two Sets of Clusters}\label{subsec.outliers-relationships}
Clusters 1--10 in the triplet distribution analysis together contain 519 genomes, a number similar to the number of genomes in clusters 1--5 for the triplet entropy analysis. Moreover, both analyses separate the 11 deliberate outliers from the 26,953 legitimate coronavirus genomes, although differently. The triplet distribution analysis places these outliers in 11 distinct clusters, while the triplet entropy degradation analysis places them all in a single cluster.

Table \ref{tab.clusterCrossTab} shows the complete and strongly block-diagonal relationship between the two sets of clusters. For clarity, cells in Table \ref{tab.clusterCrossTab} containing values of 0 are highlighted in pink. In detail,
\begin{enumerate}
\item
Triplet entropy degradation cluster 34 is, as noted above, an amalgamation of triplet distribution clusters 13--24; both contain the 11 deliberate outliers.
\item
The lone coronavirus genome in triplet distribution cluster 12 is absorbed into entropy degradation cluster 7, along with 843 other coronavirus genomes. Perhaps it is not an outlier after all.\footnote{There is further evidence to this effect in \cite{markovstructure-2021}: when clustering is done using amino acid distributions, it also ceases to be an outlier. Specifically, it is merged with cluster 11 to form a massive amino acid cluster of size 26,434.}
\item
Triplet entropy degradation clusters 6--33 disaggregate the massive, 26,433-genome triplet distribution cluster 11, modulo four additional genomes.
\item
Triplet entropy degradation clusters 1--5, containing 516 genomes and triplet distribution clusters 1--8 and 10, both containing 516 genomes, are identical collectively. These are in the upper-left corner in Table \ref{tab.clusterCrossTab}. Clearly, the two approaches are detecting the same outliers, with different nuances.
\item
Triplet distribution cluster 9, with 3 genomes, is anomalous. Each genome it contains lies in its own large entropy degradation cluster.
\end{enumerate}
Therefore, much of the scientific interpretation of outliers in \cite{markovstructure-2021}, which is based on the text string ID variable in the \NCBI\ database, carries over here.

\begin{table}[htbp]
\begin{center}
\begin{tiny}
\begin{tabular}{r|rrrrrrrrrrrrrrrrrrrrrrr|r}
\hline
Entropy & & & & & & & & & & & & & & & & & & & & & & & &
\\
Degradation & \multicolumn{23}{c|}{Triplet Distribution Cluster} &
\\
Cluster & & & & & & & & & & & & & & & & & & & & & & & &
\\
\hline
    & 1 & 2 & 3 & 4 & 5 & 6 & 7 & 8 & 9 & 10 & 11 & 12 & 13 & 14 & 15 & 16 & 17 & 18 & 19 & 20 & 21 & 22 & 23 & Sum \\
  \hline
  1 & 12 &\cellcolor{red}\textbf{0}&\cellcolor{red}\textbf{0}&\cellcolor{red}\textbf{0}&\cellcolor{red}\textbf{0}&\cellcolor{red}\textbf{0}& 137 &\cellcolor{red}\textbf{0}&\cellcolor{red}\textbf{0}&\cellcolor{red}\textbf{0}&\cellcolor{red}\textbf{0}&\cellcolor{red}\textbf{0}&\cellcolor{red}\textbf{0}&\cellcolor{red}\textbf{0}&\cellcolor{red}\textbf{0}&\cellcolor{red}\textbf{0}&\cellcolor{red}\textbf{0}&\cellcolor{red}\textbf{0}&\cellcolor{red}\textbf{0}&\cellcolor{red}\textbf{0}&\cellcolor{red}\textbf{0}&\cellcolor{red}\textbf{0}&\cellcolor{red}\textbf{0}& 149 \\
  2 &\cellcolor{red}\textbf{0}& 15 &\cellcolor{red}\textbf{0}&\cellcolor{red}\textbf{0}&\cellcolor{red}\textbf{0}&\cellcolor{red}\textbf{0}&\cellcolor{red}\textbf{0} &\cellcolor{red}\textbf{0}&\cellcolor{red}\textbf{0}& 3 &\cellcolor{red}\textbf{0}&\cellcolor{red}\textbf{0}&\cellcolor{red}\textbf{0}&\cellcolor{red}\textbf{0}&\cellcolor{red}\textbf{0}&\cellcolor{red}\textbf{0}&\cellcolor{red}\textbf{0}&\cellcolor{red}\textbf{0}&\cellcolor{red}\textbf{0}&\cellcolor{red}\textbf{0}&\cellcolor{red}\textbf{0}&\cellcolor{red}\textbf{0}&\cellcolor{red}\textbf{0}& 18 \\
  3 &\cellcolor{red}\textbf{0}&\cellcolor{red}\textbf{0}& 2 & 141 &\cellcolor{red}\textbf{0}& 27 &\cellcolor{red}\textbf{0}&\cellcolor{red}\textbf{0}&\cellcolor{red}\textbf{0}&\cellcolor{red}\textbf{0}&\cellcolor{red}\textbf{0}&\cellcolor{red}\textbf{0}&\cellcolor{red}\textbf{0}&\cellcolor{red}\textbf{0}&\cellcolor{red}\textbf{0}&\cellcolor{red}\textbf{0}&\cellcolor{red}\textbf{0}&\cellcolor{red}\textbf{0}&\cellcolor{red}\textbf{0}&\cellcolor{red}\textbf{0}&\cellcolor{red}\textbf{0}&\cellcolor{red}\textbf{0}&\cellcolor{red}\textbf{0}& 170 \\
  4 &\cellcolor{red}\textbf{0}&\cellcolor{red}\textbf{0}&\cellcolor{red}\textbf{0}&\cellcolor{red}\textbf{0}& 55 &\cellcolor{red}\textbf{0}&\cellcolor{red}\textbf{0} & 3 &\cellcolor{red}\textbf{0}&\cellcolor{red}\textbf{0}&\cellcolor{red}\textbf{0}&\cellcolor{red}\textbf{0}&\cellcolor{red}\textbf{0}&\cellcolor{red}\textbf{0}&\cellcolor{red}\textbf{0}&\cellcolor{red}\textbf{0}&\cellcolor{red}\textbf{0}&\cellcolor{red}\textbf{0}&\cellcolor{red}\textbf{0}&\cellcolor{red}\textbf{0}&\cellcolor{red}\textbf{0}&\cellcolor{red}\textbf{0}&\cellcolor{red}\textbf{0}& 58 \\
  5 &\cellcolor{red}\textbf{0}&\cellcolor{red}\textbf{0}&\cellcolor{red}\textbf{0}&\cellcolor{red}\textbf{0}&\cellcolor{red}\textbf{0}&\cellcolor{red}\textbf{0}& 121 &\cellcolor{red}\textbf{0}&\cellcolor{red}\textbf{0}&\cellcolor{red}\textbf{0}&\cellcolor{red}\textbf{0}&\cellcolor{red}\textbf{0}&\cellcolor{red}\textbf{0}&\cellcolor{red}\textbf{0}&\cellcolor{red}\textbf{0}&\cellcolor{red}\textbf{0}&\cellcolor{red}\textbf{0}&\cellcolor{red}\textbf{0}&\cellcolor{red}\textbf{0}&\cellcolor{red}\textbf{0}&\cellcolor{red}\textbf{0}&\cellcolor{red}\textbf{0}&\cellcolor{red}\textbf{0}& 121 \\
  6 &\cellcolor{red}\textbf{0}&\cellcolor{red}\textbf{0}&\cellcolor{red}\textbf{0}&\cellcolor{red}\textbf{0}&\cellcolor{red}\textbf{0}&\cellcolor{red}\textbf{0}&\cellcolor{red}\textbf{0}&\cellcolor{red}\textbf{0}& 1 &\cellcolor{red}\textbf{0}& 354 &\cellcolor{red}\textbf{0}&\cellcolor{red}\textbf{0}&\cellcolor{red}\textbf{0}&\cellcolor{red}\textbf{0}&\cellcolor{red}\textbf{0} &\cellcolor{red}\textbf{0}&\cellcolor{red}\textbf{0}&\cellcolor{red}\textbf{0}&\cellcolor{red}\textbf{0}&\cellcolor{red}\textbf{0}&\cellcolor{red}\textbf{0}&\cellcolor{red}\textbf{0}& 355 \\
  7 &\cellcolor{red}\textbf{0}&\cellcolor{red}\textbf{0}&\cellcolor{red}\textbf{0}&\cellcolor{red}\textbf{0}&\cellcolor{red}\textbf{0}&\cellcolor{red}\textbf{0}&\cellcolor{red}\textbf{0}&\cellcolor{red}\textbf{0}& 1 &\cellcolor{red}\textbf{0}& 842 & 1 &\cellcolor{red}\textbf{0}&\cellcolor{red}\textbf{0}&\cellcolor{red}\textbf{0}&\cellcolor{red}\textbf{0}&\cellcolor{red}\textbf{0}&\cellcolor{red}\textbf{0}&\cellcolor{red}\textbf{0}&\cellcolor{red}\textbf{0}&\cellcolor{red}\textbf{0}&\cellcolor{red}\textbf{0}&\cellcolor{red}\textbf{0}& 844 \\
  8 &\cellcolor{red}\textbf{0}&\cellcolor{red}\textbf{0}&\cellcolor{red}\textbf{0}&\cellcolor{red}\textbf{0}&\cellcolor{red}\textbf{0}&\cellcolor{red}\textbf{0}&\cellcolor{red}\textbf{0}&\cellcolor{red}\textbf{0}& 1 &\cellcolor{red}\textbf{0}& 313 &\cellcolor{red}\textbf{0}&\cellcolor{red}\textbf{0}&\cellcolor{red}\textbf{0}&\cellcolor{red}\textbf{0}&\cellcolor{red}\textbf{0} &\cellcolor{red}\textbf{0}&\cellcolor{red}\textbf{0}&\cellcolor{red}\textbf{0}&\cellcolor{red}\textbf{0}&\cellcolor{red}\textbf{0}&\cellcolor{red}\textbf{0}&\cellcolor{red}\textbf{0}& 314 \\
  9 &\cellcolor{red}\textbf{0}&\cellcolor{red}\textbf{0}&\cellcolor{red}\textbf{0}&\cellcolor{red}\textbf{0}&\cellcolor{red}\textbf{0}&\cellcolor{red}\textbf{0}&\cellcolor{red}\textbf{0}&\cellcolor{red}\textbf{0}&\cellcolor{red}\textbf{0}&\cellcolor{red}\textbf{0}& 1470 &\cellcolor{red}\textbf{0}&\cellcolor{red}\textbf{0}&\cellcolor{red}\textbf{0}&\cellcolor{red}\textbf{0}&\cellcolor{red}\textbf{0}&\cellcolor{red}\textbf{0}&\cellcolor{red}\textbf{0}&\cellcolor{red}\textbf{0}&\cellcolor{red}\textbf{0}&\cellcolor{red}\textbf{0}&\cellcolor{red}\textbf{0}&\cellcolor{red}\textbf{0}& 1470 \\
  10 &\cellcolor{red}\textbf{0}&\cellcolor{red}\textbf{0}&\cellcolor{red}\textbf{0}&\cellcolor{red}\textbf{0}&\cellcolor{red}\textbf{0}&\cellcolor{red}\textbf{0}&\cellcolor{red}\textbf{0}&\cellcolor{red}\textbf{0}&\cellcolor{red}\textbf{0}&\cellcolor{red}\textbf{0}& 1466 &\cellcolor{red}\textbf{0}&\cellcolor{red}\textbf{0}&\cellcolor{red}\textbf{0}&\cellcolor{red}\textbf{0}&\cellcolor{red}\textbf{0}&\cellcolor{red}\textbf{0}&\cellcolor{red}\textbf{0}&\cellcolor{red}\textbf{0}&\cellcolor{red}\textbf{0}&\cellcolor{red}\textbf{0}&\cellcolor{red}\textbf{0}&\cellcolor{red}\textbf{0}& 1466 \\
  11 &\cellcolor{red}\textbf{0}&\cellcolor{red}\textbf{0}&\cellcolor{red}\textbf{0}&\cellcolor{red}\textbf{0}&\cellcolor{red}\textbf{0}&\cellcolor{red}\textbf{0}&\cellcolor{red}\textbf{0}&\cellcolor{red}\textbf{0}&\cellcolor{red}\textbf{0}&\cellcolor{red}\textbf{0}& 1005 &\cellcolor{red}\textbf{0}&\cellcolor{red}\textbf{0}&\cellcolor{red}\textbf{0}&\cellcolor{red}\textbf{0}&\cellcolor{red}\textbf{0}&\cellcolor{red}\textbf{0}&\cellcolor{red}\textbf{0}&\cellcolor{red}\textbf{0}&\cellcolor{red}\textbf{0}&\cellcolor{red}\textbf{0}&\cellcolor{red}\textbf{0}&\cellcolor{red}\textbf{0}& 1005 \\
  12 &\cellcolor{red}\textbf{0}&\cellcolor{red}\textbf{0}&\cellcolor{red}\textbf{0}&\cellcolor{red}\textbf{0}&\cellcolor{red}\textbf{0}&\cellcolor{red}\textbf{0}&\cellcolor{red}\textbf{0}&\cellcolor{red}\textbf{0}&\cellcolor{red}\textbf{0}&\cellcolor{red}\textbf{0}& 923 &\cellcolor{red}\textbf{0}&\cellcolor{red}\textbf{0}&\cellcolor{red}\textbf{0}&\cellcolor{red}\textbf{0}&\cellcolor{red}\textbf{0}&\cellcolor{red}\textbf{0}&\cellcolor{red}\textbf{0}&\cellcolor{red}\textbf{0}&\cellcolor{red}\textbf{0}&\cellcolor{red}\textbf{0}&\cellcolor{red}\textbf{0}&\cellcolor{red}\textbf{0}& 923 \\
  13 &\cellcolor{red}\textbf{0}&\cellcolor{red}\textbf{0}&\cellcolor{red}\textbf{0}&\cellcolor{red}\textbf{0}&\cellcolor{red}\textbf{0}&\cellcolor{red}\textbf{0}&\cellcolor{red}\textbf{0}&\cellcolor{red}\textbf{0}&\cellcolor{red}\textbf{0}&\cellcolor{red}\textbf{0}& 1215 &\cellcolor{red}\textbf{0}&\cellcolor{red}\textbf{0}&\cellcolor{red}\textbf{0}&\cellcolor{red}\textbf{0}&\cellcolor{red}\textbf{0}&\cellcolor{red}\textbf{0}&\cellcolor{red}\textbf{0}&\cellcolor{red}\textbf{0}&\cellcolor{red}\textbf{0}&\cellcolor{red}\textbf{0}&\cellcolor{red}\textbf{0}&\cellcolor{red}\textbf{0}& 1215 \\
  14 &\cellcolor{red}\textbf{0}&\cellcolor{red}\textbf{0}&\cellcolor{red}\textbf{0}&\cellcolor{red}\textbf{0}&\cellcolor{red}\textbf{0}&\cellcolor{red}\textbf{0}&\cellcolor{red}\textbf{0}&\cellcolor{red}\textbf{0}&\cellcolor{red}\textbf{0}&\cellcolor{red}\textbf{0}& 849 &\cellcolor{red}\textbf{0}&\cellcolor{red}\textbf{0}&\cellcolor{red}\textbf{0}&\cellcolor{red}\textbf{0}&\cellcolor{red}\textbf{0}&\cellcolor{red}\textbf{0}&\cellcolor{red}\textbf{0}&\cellcolor{red}\textbf{0}&\cellcolor{red}\textbf{0}&\cellcolor{red}\textbf{0}&\cellcolor{red}\textbf{0}&\cellcolor{red}\textbf{0}& 849 \\
  15 &\cellcolor{red}\textbf{0}&\cellcolor{red}\textbf{0}&\cellcolor{red}\textbf{0}&\cellcolor{red}\textbf{0}&\cellcolor{red}\textbf{0}&\cellcolor{red}\textbf{0}&\cellcolor{red}\textbf{0}&\cellcolor{red}\textbf{0}&\cellcolor{red}\textbf{0}&\cellcolor{red}\textbf{0}& 858 &\cellcolor{red}\textbf{0}&\cellcolor{red}\textbf{0}&\cellcolor{red}\textbf{0}&\cellcolor{red}\textbf{0}&\cellcolor{red}\textbf{0}&\cellcolor{red}\textbf{0}&\cellcolor{red}\textbf{0}&\cellcolor{red}\textbf{0}&\cellcolor{red}\textbf{0}&\cellcolor{red}\textbf{0}&\cellcolor{red}\textbf{0}&\cellcolor{red}\textbf{0}& 858 \\
  16 &\cellcolor{red}\textbf{0}&\cellcolor{red}\textbf{0}&\cellcolor{red}\textbf{0}&\cellcolor{red}\textbf{0}&\cellcolor{red}\textbf{0}&\cellcolor{red}\textbf{0}&\cellcolor{red}\textbf{0}&\cellcolor{red}\textbf{0}&\cellcolor{red}\textbf{0}&\cellcolor{red}\textbf{0}& 526 &\cellcolor{red}\textbf{0}&\cellcolor{red}\textbf{0}&\cellcolor{red}\textbf{0}&\cellcolor{red}\textbf{0}&\cellcolor{red}\textbf{0}&\cellcolor{red}\textbf{0}&\cellcolor{red}\textbf{0}&\cellcolor{red}\textbf{0}&\cellcolor{red}\textbf{0}&\cellcolor{red}\textbf{0}&\cellcolor{red}\textbf{0}&\cellcolor{red}\textbf{0}& 526 \\
  17 &\cellcolor{red}\textbf{0}&\cellcolor{red}\textbf{0}&\cellcolor{red}\textbf{0}&\cellcolor{red}\textbf{0}&\cellcolor{red}\textbf{0}&\cellcolor{red}\textbf{0}&\cellcolor{red}\textbf{0}&\cellcolor{red}\textbf{0}&\cellcolor{red}\textbf{0}&\cellcolor{red}\textbf{0}& 1138 &\cellcolor{red}\textbf{0}&\cellcolor{red}\textbf{0}&\cellcolor{red}\textbf{0}&\cellcolor{red}\textbf{0}&\cellcolor{red}\textbf{0}&\cellcolor{red}\textbf{0}&\cellcolor{red}\textbf{0}&\cellcolor{red}\textbf{0}&\cellcolor{red}\textbf{0}&\cellcolor{red}\textbf{0}&\cellcolor{red}\textbf{0}&\cellcolor{red}\textbf{0}& 1138 \\
  18 &\cellcolor{red}\textbf{0}&\cellcolor{red}\textbf{0}&\cellcolor{red}\textbf{0}&\cellcolor{red}\textbf{0}&\cellcolor{red}\textbf{0}&\cellcolor{red}\textbf{0}&\cellcolor{red}\textbf{0}&\cellcolor{red}\textbf{0}&\cellcolor{red}\textbf{0}&\cellcolor{red}\textbf{0}& 946 &\cellcolor{red}\textbf{0}&\cellcolor{red}\textbf{0}&\cellcolor{red}\textbf{0}&\cellcolor{red}\textbf{0}&\cellcolor{red}\textbf{0}&\cellcolor{red}\textbf{0}&\cellcolor{red}\textbf{0}&\cellcolor{red}\textbf{0}&\cellcolor{red}\textbf{0}&\cellcolor{red}\textbf{0}&\cellcolor{red}\textbf{0}&\cellcolor{red}\textbf{0}& 946 \\
  19 &\cellcolor{red}\textbf{0}&\cellcolor{red}\textbf{0}&\cellcolor{red}\textbf{0}&\cellcolor{red}\textbf{0}&\cellcolor{red}\textbf{0}&\cellcolor{red}\textbf{0}&\cellcolor{red}\textbf{0}&\cellcolor{red}\textbf{0}&\cellcolor{red}\textbf{0}&\cellcolor{red}\textbf{0}& 1259 &\cellcolor{red}\textbf{0}&\cellcolor{red}\textbf{0}&\cellcolor{red}\textbf{0}&\cellcolor{red}\textbf{0}&\cellcolor{red}\textbf{0}&\cellcolor{red}\textbf{0}&\cellcolor{red}\textbf{0}&\cellcolor{red}\textbf{0}&\cellcolor{red}\textbf{0}&\cellcolor{red}\textbf{0}&\cellcolor{red}\textbf{0}&\cellcolor{red}\textbf{0}& 1259 \\
  20 &\cellcolor{red}\textbf{0}&\cellcolor{red}\textbf{0}&\cellcolor{red}\textbf{0}&\cellcolor{red}\textbf{0}&\cellcolor{red}\textbf{0}&\cellcolor{red}\textbf{0}&\cellcolor{red}\textbf{0}&\cellcolor{red}\textbf{0}&\cellcolor{red}\textbf{0}&\cellcolor{red}\textbf{0}& 1111 &\cellcolor{red}\textbf{0}&\cellcolor{red}\textbf{0}&\cellcolor{red}\textbf{0}&\cellcolor{red}\textbf{0}&\cellcolor{red}\textbf{0}&\cellcolor{red}\textbf{0}&\cellcolor{red}\textbf{0}&\cellcolor{red}\textbf{0}&\cellcolor{red}\textbf{0}&\cellcolor{red}\textbf{0}&\cellcolor{red}\textbf{0}&\cellcolor{red}\textbf{0}& 1111 \\
  21 &\cellcolor{red}\textbf{0}&\cellcolor{red}\textbf{0}&\cellcolor{red}\textbf{0}&\cellcolor{red}\textbf{0}&\cellcolor{red}\textbf{0}&\cellcolor{red}\textbf{0}&\cellcolor{red}\textbf{0}&\cellcolor{red}\textbf{0}&\cellcolor{red}\textbf{0}&\cellcolor{red}\textbf{0}& 1393 &\cellcolor{red}\textbf{0}&\cellcolor{red}\textbf{0}&\cellcolor{red}\textbf{0}&\cellcolor{red}\textbf{0}&\cellcolor{red}\textbf{0}&\cellcolor{red}\textbf{0}&\cellcolor{red}\textbf{0}&\cellcolor{red}\textbf{0}&\cellcolor{red}\textbf{0}&\cellcolor{red}\textbf{0}&\cellcolor{red}\textbf{0}&\cellcolor{red}\textbf{0}& 1393 \\
  22 &\cellcolor{red}\textbf{0}&\cellcolor{red}\textbf{0}&\cellcolor{red}\textbf{0}&\cellcolor{red}\textbf{0}&\cellcolor{red}\textbf{0}&\cellcolor{red}\textbf{0}&\cellcolor{red}\textbf{0}&\cellcolor{red}\textbf{0}&\cellcolor{red}\textbf{0}&\cellcolor{red}\textbf{0}& 776 &\cellcolor{red}\textbf{0}&\cellcolor{red}\textbf{0}&\cellcolor{red}\textbf{0}&\cellcolor{red}\textbf{0}&\cellcolor{red}\textbf{0}&\cellcolor{red}\textbf{0}&\cellcolor{red}\textbf{0}&\cellcolor{red}\textbf{0}&\cellcolor{red}\textbf{0}&\cellcolor{red}\textbf{0}&\cellcolor{red}\textbf{0}&\cellcolor{red}\textbf{0}& 776 \\
  23 &\cellcolor{red}\textbf{0}&\cellcolor{red}\textbf{0}&\cellcolor{red}\textbf{0}&\cellcolor{red}\textbf{0}&\cellcolor{red}\textbf{0}&\cellcolor{red}\textbf{0}&\cellcolor{red}\textbf{0}&\cellcolor{red}\textbf{0}&\cellcolor{red}\textbf{0}&\cellcolor{red}\textbf{0}& 1027 &\cellcolor{red}\textbf{0}&\cellcolor{red}\textbf{0}&\cellcolor{red}\textbf{0}&\cellcolor{red}\textbf{0}&\cellcolor{red}\textbf{0}&\cellcolor{red}\textbf{0}&\cellcolor{red}\textbf{0}&\cellcolor{red}\textbf{0}&\cellcolor{red}\textbf{0}&\cellcolor{red}\textbf{0}&\cellcolor{red}\textbf{0}&\cellcolor{red}\textbf{0}& 1027 \\
  24 &\cellcolor{red}\textbf{0}&\cellcolor{red}\textbf{0}&\cellcolor{red}\textbf{0}&\cellcolor{red}\textbf{0}&\cellcolor{red}\textbf{0}&\cellcolor{red}\textbf{0}&\cellcolor{red}\textbf{0}&\cellcolor{red}\textbf{0}&\cellcolor{red}\textbf{0}&\cellcolor{red}\textbf{0}& 1176 &\cellcolor{red}\textbf{0}&\cellcolor{red}\textbf{0}&\cellcolor{red}\textbf{0}&\cellcolor{red}\textbf{0}&\cellcolor{red}\textbf{0}&\cellcolor{red}\textbf{0}&\cellcolor{red}\textbf{0}&\cellcolor{red}\textbf{0}&\cellcolor{red}\textbf{0}&\cellcolor{red}\textbf{0}&\cellcolor{red}\textbf{0}&\cellcolor{red}\textbf{0}& 1176 \\
  25 &\cellcolor{red}\textbf{0}&\cellcolor{red}\textbf{0}&\cellcolor{red}\textbf{0}&\cellcolor{red}\textbf{0}&\cellcolor{red}\textbf{0}&\cellcolor{red}\textbf{0}&\cellcolor{red}\textbf{0}&\cellcolor{red}\textbf{0}&\cellcolor{red}\textbf{0}&\cellcolor{red}\textbf{0}& 825 &\cellcolor{red}\textbf{0}&\cellcolor{red}\textbf{0}&\cellcolor{red}\textbf{0}&\cellcolor{red}\textbf{0}&\cellcolor{red}\textbf{0}&\cellcolor{red}\textbf{0}&\cellcolor{red}\textbf{0}&\cellcolor{red}\textbf{0}&\cellcolor{red}\textbf{0}&\cellcolor{red}\textbf{0}&\cellcolor{red}\textbf{0}&\cellcolor{red}\textbf{0}& 825 \\
  26 &\cellcolor{red}\textbf{0}&\cellcolor{red}\textbf{0}&\cellcolor{red}\textbf{0}&\cellcolor{red}\textbf{0}&\cellcolor{red}\textbf{0}&\cellcolor{red}\textbf{0}&\cellcolor{red}\textbf{0}&\cellcolor{red}\textbf{0}&\cellcolor{red}\textbf{0}&\cellcolor{red}\textbf{0}& 1115 &\cellcolor{red}\textbf{0}&\cellcolor{red}\textbf{0}&\cellcolor{red}\textbf{0}&\cellcolor{red}\textbf{0}&\cellcolor{red}\textbf{0}&\cellcolor{red}\textbf{0}&\cellcolor{red}\textbf{0}&\cellcolor{red}\textbf{0}&\cellcolor{red}\textbf{0}&\cellcolor{red}\textbf{0}&\cellcolor{red}\textbf{0}&\cellcolor{red}\textbf{0}& 1115 \\
  27 &\cellcolor{red}\textbf{0}&\cellcolor{red}\textbf{0}&\cellcolor{red}\textbf{0}&\cellcolor{red}\textbf{0}&\cellcolor{red}\textbf{0}&\cellcolor{red}\textbf{0}&\cellcolor{red}\textbf{0}&\cellcolor{red}\textbf{0}&\cellcolor{red}\textbf{0}&\cellcolor{red}\textbf{0}& 1069 &\cellcolor{red}\textbf{0}&\cellcolor{red}\textbf{0}&\cellcolor{red}\textbf{0}&\cellcolor{red}\textbf{0}&\cellcolor{red}\textbf{0}&\cellcolor{red}\textbf{0}&\cellcolor{red}\textbf{0}&\cellcolor{red}\textbf{0}&\cellcolor{red}\textbf{0}&\cellcolor{red}\textbf{0}&\cellcolor{red}\textbf{0}&\cellcolor{red}\textbf{0}& 1069 \\
  28 &\cellcolor{red}\textbf{0}&\cellcolor{red}\textbf{0}&\cellcolor{red}\textbf{0}&\cellcolor{red}\textbf{0}&\cellcolor{red}\textbf{0}&\cellcolor{red}\textbf{0}&\cellcolor{red}\textbf{0}&\cellcolor{red}\textbf{0}&\cellcolor{red}\textbf{0}&\cellcolor{red}\textbf{0}& 467 &\cellcolor{red}\textbf{0}&\cellcolor{red}\textbf{0}&\cellcolor{red}\textbf{0}&\cellcolor{red}\textbf{0}&\cellcolor{red}\textbf{0}&\cellcolor{red}\textbf{0}&\cellcolor{red}\textbf{0}&\cellcolor{red}\textbf{0}&\cellcolor{red}\textbf{0}&\cellcolor{red}\textbf{0}&\cellcolor{red}\textbf{0}&\cellcolor{red}\textbf{0}& 467 \\
  29 &\cellcolor{red}\textbf{0}&\cellcolor{red}\textbf{0}&\cellcolor{red}\textbf{0}&\cellcolor{red}\textbf{0}&\cellcolor{red}\textbf{0}&\cellcolor{red}\textbf{0}&\cellcolor{red}\textbf{0}&\cellcolor{red}\textbf{0}&\cellcolor{red}\textbf{0}&\cellcolor{red}\textbf{0}& 949 &\cellcolor{red}\textbf{0}&\cellcolor{red}\textbf{0}&\cellcolor{red}\textbf{0}&\cellcolor{red}\textbf{0}&\cellcolor{red}\textbf{0}&\cellcolor{red}\textbf{0}&\cellcolor{red}\textbf{0}&\cellcolor{red}\textbf{0}&\cellcolor{red}\textbf{0}&\cellcolor{red}\textbf{0}&\cellcolor{red}\textbf{0}&\cellcolor{red}\textbf{0}& 949 \\
  30 &\cellcolor{red}\textbf{0}&\cellcolor{red}\textbf{0}&\cellcolor{red}\textbf{0}&\cellcolor{red}\textbf{0}&\cellcolor{red}\textbf{0}&\cellcolor{red}\textbf{0}&\cellcolor{red}\textbf{0}&\cellcolor{red}\textbf{0}&\cellcolor{red}\textbf{0}&\cellcolor{red}\textbf{0}& 1261 &\cellcolor{red}\textbf{0}&\cellcolor{red}\textbf{0}&\cellcolor{red}\textbf{0}&\cellcolor{red}\textbf{0}&\cellcolor{red}\textbf{0}&\cellcolor{red}\textbf{0}&\cellcolor{red}\textbf{0}&\cellcolor{red}\textbf{0}&\cellcolor{red}\textbf{0}&\cellcolor{red}\textbf{0}&\cellcolor{red}\textbf{0}&\cellcolor{red}\textbf{0}& 1261 \\
  31 &\cellcolor{red}\textbf{0}&\cellcolor{red}\textbf{0}&\cellcolor{red}\textbf{0}&\cellcolor{red}\textbf{0}&\cellcolor{red}\textbf{0}&\cellcolor{red}\textbf{0}&\cellcolor{red}\textbf{0}&\cellcolor{red}\textbf{0}&\cellcolor{red}\textbf{0}&\cellcolor{red}\textbf{0}& 928 &\cellcolor{red}\textbf{0}&\cellcolor{red}\textbf{0}&\cellcolor{red}\textbf{0}&\cellcolor{red}\textbf{0}&\cellcolor{red}\textbf{0}&\cellcolor{red}\textbf{0}&\cellcolor{red}\textbf{0}&\cellcolor{red}\textbf{0}&\cellcolor{red}\textbf{0}&\cellcolor{red}\textbf{0}&\cellcolor{red}\textbf{0}&\cellcolor{red}\textbf{0}& 928 \\
  32 &\cellcolor{red}\textbf{0}&\cellcolor{red}\textbf{0}&\cellcolor{red}\textbf{0}&\cellcolor{red}\textbf{0}&\cellcolor{red}\textbf{0}&\cellcolor{red}\textbf{0}&\cellcolor{red}\textbf{0}&\cellcolor{red}\textbf{0}&\cellcolor{red}\textbf{0}&\cellcolor{red}\textbf{0}& 589 &\cellcolor{red}\textbf{0}&\cellcolor{red}\textbf{0}&\cellcolor{red}\textbf{0}&\cellcolor{red}\textbf{0}&\cellcolor{red}\textbf{0}&\cellcolor{red}\textbf{0}&\cellcolor{red}\textbf{0}&\cellcolor{red}\textbf{0}&\cellcolor{red}\textbf{0}&\cellcolor{red}\textbf{0}&\cellcolor{red}\textbf{0}&\cellcolor{red}\textbf{0}& 589 \\
  33 &\cellcolor{red}\textbf{0}&\cellcolor{red}\textbf{0}&\cellcolor{red}\textbf{0}&\cellcolor{red}\textbf{0}&\cellcolor{red}\textbf{0}&\cellcolor{red}\textbf{0}&\cellcolor{red}\textbf{0}&\cellcolor{red}\textbf{0}&\cellcolor{red}\textbf{0}&\cellcolor{red}\textbf{0}& 583 &\cellcolor{red}\textbf{0}&\cellcolor{red}\textbf{0}&\cellcolor{red}\textbf{0}&\cellcolor{red}\textbf{0}&\cellcolor{red}\textbf{0}&\cellcolor{red}\textbf{0}&\cellcolor{red}\textbf{0}&\cellcolor{red}\textbf{0}&\cellcolor{red}\textbf{0}&\cellcolor{red}\textbf{0}&\cellcolor{red}\textbf{0}&\cellcolor{red}\textbf{0}& 583 \\
  34 &\cellcolor{red}\textbf{0}&\cellcolor{red}\textbf{0}&\cellcolor{red}\textbf{0}&\cellcolor{red}\textbf{0}&\cellcolor{red}\textbf{0}&\cellcolor{red}\textbf{0}&\cellcolor{red}\textbf{0}&\cellcolor{red}\textbf{0}&\cellcolor{red}\textbf{0}&\cellcolor{red}\textbf{0}&\cellcolor{red}\textbf{0}&\cellcolor{red}\textbf{0}& 1 & 1 & 1 & 1 & 1 & 1 & 1 & 1 & 1 & 1 & 1 & 11 \\
  \hline
  Sum & 12 & 15 & 2 & 141 & 55 & 27 & 258 & 3 & 3 & 3 & 26433 & 1 & 1 & 1 & 1 & 1 & 1 & 1 & 1 & 1 & 1 & 1 & 1 & 26964 \\
   \hline
\end{tabular}
\end{tiny}
\caption{Cross-tabulation of the 34 degradation clusters (rows) and the 23 triplet distribution clusters (columns). Cells containing zeros are shaded in pink.}
\label{tab.clusterCrossTab}
\end{center}
\end{table}

\section{Extensions}\label{sec.extensions}
This section extends \S\ref{sec.bases} to higher-level DNA structure---specifically, repeats and palindromes (\S\ref{subsec.dnastructure}), and to genomes other than viruses (\S\ref{subsec.generalizability}).

\subsection{Higher-Order DNA Structure}\label{subsec.dnastructure}
We showed in \S \ref{sec.bases} that degradation attenuates (relatively) low-dimensional genome characteristics such as tuple distributions (Figure \ref{fig.degradedbasedistributions}). We see here that more complex structure such as repeats and palindromes is affected even more strongly. As exemplar, we use an \emph{E. coli} genome of length 4,639,675 downloaded from \NCBI; the same genome appears again in Figure \ref{fig.entropydegradation-other}.

\subsubsection{Repeats}\label{subsubsec.repeats}
Repeats are inherent to non-virus genomes, leading, \emph{inter alia}, to the discovery of \CRISPR\ in \emph{E. coli} \citep{mojica-crispr-2000}. Table \ref{tab.Ecoli-repeats} shows the effect of \MV\ degradation on the numbers of repeats of various lengths in the \emph{E. coli} genome. The column for length 29, rather than 30, honors the original discovery of \CRISPR. By 500 \MV\ iterations, all repeats of length 20 or longer have been obliterated. Those of length 29 are gone at 300 iterations.

\begin{table}[htbp]
\begin{center}\begin{tabular}{rrrrrr}
  \hline
  & \multicolumn{5}{c}{Repeat Length} \\
Iteration & 20 & 25 & 29 & 35 & 40 \\
  \hline
0 & 37285 & 35588 & 34667 & 33429 & 32530 \\
  100 & 2865 & 1094 & 487 & 153 & 58 \\
  200 & 640 & 156 & 47 & 11 & 2 \\
  300 & 152 & 11 & 0 & 0 & 0 \\
  400 & 36 & 1 & 0 & 0 & 0 \\
  500 & 17 & 0 & 0 & 0 & 0 \\
  600 & 14 & 0 & 0 & 0 & 0 \\
  700 & 6 & 0 & 0 & 0 & 0 \\
  800 & 8 & 0 & 0 & 0 & 0 \\
  900 & 10 & 0 & 0 & 0 & 0 \\
  1000 & 6 & 0 & 0 & 0 & 0 \\
   \hline
\end{tabular}\end{center}
\caption{Numbers of repeats of length 20, 25, 29, 35 and 40 (columns) in \emph{E. coli} as a function of \MV\ iterations (rows).}
\label{tab.Ecoli-repeats}
\end{table}

\subsubsection{Palindromes}\label{subsubsec.palindromes}
Genomic palindromes, unlike those in ordinary language, consist of a sequence of bases followed immediately by its reverse complement. So, an example is ATTCGATT||AATCGAAT.\footnote{The || has been inserted for visual clarity.} In what follows, palindromes are parameterized by half-length; the example has half-length 8. Their behavior with respect to \MV\ degradation differs somewhat from that of repeats.

Table \ref{tab.Ecoli-palindromes} shows that long palindromes (half-lengths 12, 14 and 16), are not plentiful to begin with, and vanish, modulo noise discussed momentarily, within 100 \MV\ iterations. Palindromes with half-length 8 and 10 do decline in number, but do not vanish, even by 2000 iterations. Moreover, their numbers can increase, although not enormously. Palindromes of half length 6 barely diminish at all, and fluctuate substantially, The noise and the increases suggest that short palindromes differ from the other genome features discussed in this paper, and especially from repeats. They are resistant to the \MV\ degradation, and can even be produced by it. This is not surprising because very short repeats are low-dimensional and may, therefore, be too short to be interesting biologically.

\begin{table}[htbp]
\begin{center}
\begin{tabular}{rrrrrrr}
  \hline
  & \multicolumn{6}{c}{Half-Length}
\\
Iterations & 6 & 8 & 10 & 12 & 14 & 16 \\
  \hline
  0 & 1128 & 113 & 22 & 11 & 2 & 1 \\
  100 & 1149 & 102 & 7 & 1 & 0 & 0 \\
  200 & 1163 & 90 & 4 & 1 & 0 & 0 \\
  300 & 1209 & 87 & 6 & 1 & 0 & 0 \\
  400 & 1137 & 81 & 6 & 1 & 0 & 0 \\
  500 & 1114 & 79 & 5 & 0 & 0 & 0 \\
  600 & 1141 & 79 & 8 & 0 & 0 & 0 \\
  700 & 1126 & 81 & 8 & 0 & 0 & 0 \\
  800 & 1130 & 75 & 4 & 1 & 1 & 0 \\
  900 & 1140 & 67 & 3 & 1 & 1 & 0 \\
  1000 & 1077 & 62 & 4 & 0 & 0 & 0 \\
  1500 & 1104 & 66 & 2 & 0 & 0 & 0 \\
  2000 & 1072 & 68 & 3 & 0 & 0 & 0 \\
   \hline
\end{tabular}
\end{center}
\caption{Numbers of palindromes of half lengths 6, 8, 10, 12, 14 and 16 in \emph{E. coli} (columns) as a function of \MV\ iterations (rows).}
\label{tab.Ecoli-palindromes}
\end{table}

\subsection{Other Genomes}\label{subsec.generalizability}
\begin{figure}[htbp]
\begin{center}
\includegraphics[width=4in]{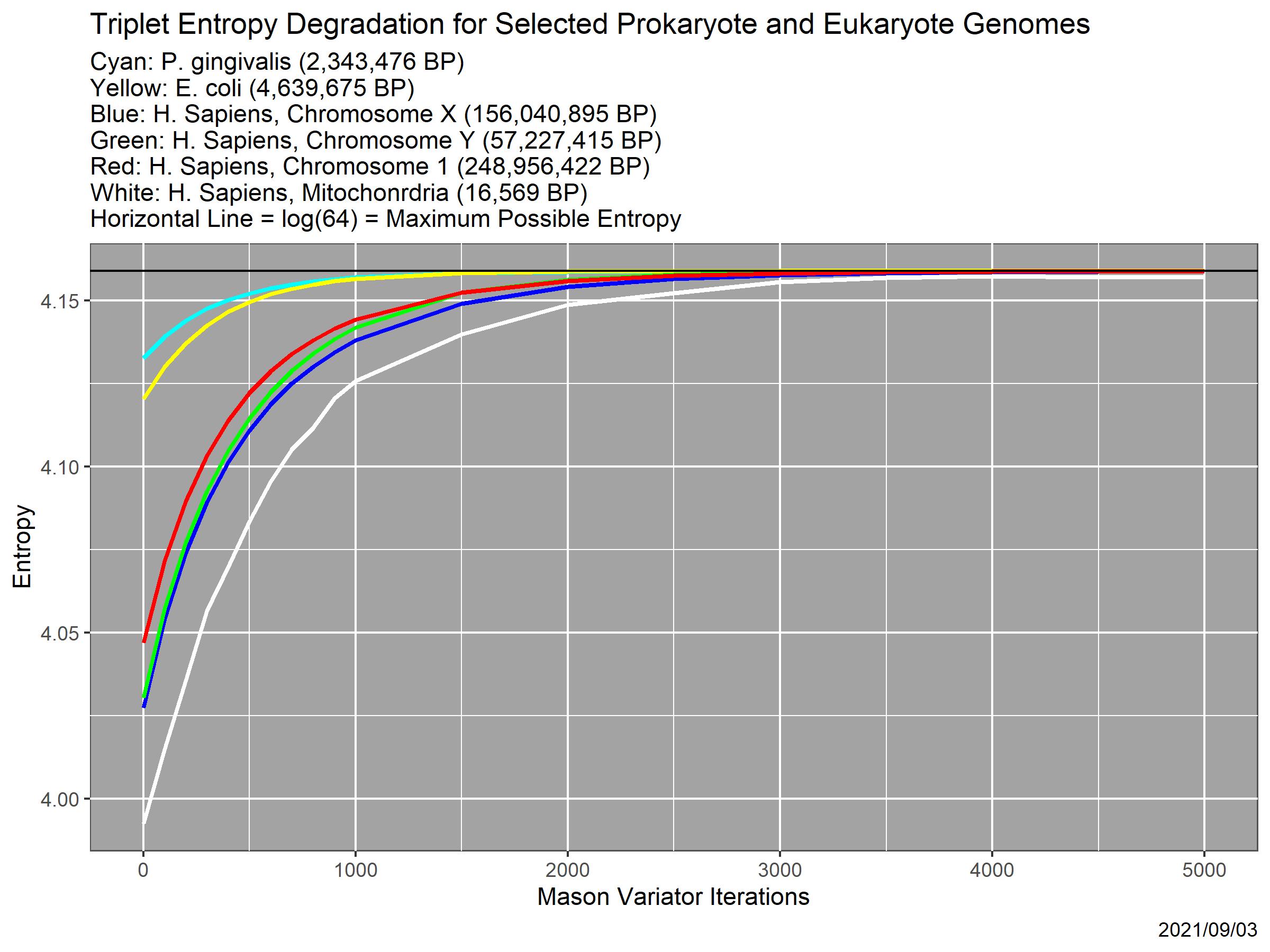}
\end{center}
\caption{Entropy as a function of \MV\ iterations, for selected prokaryote and eukaryote genomes.}
\label{fig.entropydegradation-other}
\end{figure}

Figure \ref{fig.entropydegradation-other} demonstrates that degradation behavior is not confined to viruses. It is the analog of Figure \ref{fig.entropydegradation-26964} for two bacterial genomes---\emph{P. gingivalis} and \emph{E. coli} (the same genome in \S\ref{subsec.dnastructure}), for three human chromosomes---1, X and Y, and for human mitochondrial DNA. All six genomes were downloaded from \NCBI; the human genome is identified as GRCh38.

\section{Conclusions}\label{sec.conclusions}
In this paper, we have introduced and investigated a new, degradation-based approach to data quality for genome sequence databases, and established that it is sound scientifically and statistically. Our principal application is to outlier detection, and our methods are demonstrably effective.

At least three paths for further research are clear. The first is that our paradigm, at this point, does not produce quantified uncertainties about the decisions it may engender, so tools such as Specified Certainty Classification \citep{scc-2021} cannot be applied. Following this path also requires, as raised in Appendix \ref{app.dq}, more explicit attention to the decisions to be made using the data. To consider decision quality fully leads to the second path---better understanding of the effects of data quality on bioinformatics software pipelines. Now that we can create data of demonstrably and quantifiably lower quality, this path is feasible. Third and more speculatively, there is the relationship between data quality and adversarial attacks on genome databases or software pipelines \citep{wildpatterns-2018, farbiash-puzis-2020, biodefense-2019}. Attempts to ``pollute'' databases with (what may turn out to be) low quality genomes are potentially detectable using the outlier identification strategies presented here. Risk-utility paradigms for statistical disclosure limitation (SDL) discussed in \cite{isr-2011} are relevant, especially the need to distinguish attackers from legitimate users of the data.

Finally, there is clear potential to extend our paradigm to contexts other than genomics, provided that a credible generative model for quality degradation can be constructed. To illustrate, in the official statistics context of Appendix \ref{app.dq}, one simply needs a mechanism, such as the microsimulation tool in \cite{wssm201210}, to simulate one of more forms of total survey error.

\section*{Acknowledgements}
This research was supported in part by NIH grant 5R01AI100947--06, ``Algorithms and Software for the Assembly of Metagenomic Data,'' to the University of Maryland College Park (Mihai Pop, PI).

\appendix

\section{Background on Data Quality}\label{app.dq}
This section draws on \cite{dq-statmeth06}, with additional material from \OMB\ documents \citep{omb-standards-2006}.

Data quality is a complex, multi-dimensional construct driven ultimately by data usage and subsequent decisions \citep{dqreport, dq-statmeth06, karr-josdiscussion-2013}. There are two key questions: What are the intended uses of the data? What decisions are to be made based in part on the data? In genomics, quality often focuses solely on errors. But, there are multiple hyperdimensions of data quality, each containing multiple dimensions:
\begin{description}
\item[Process]
Dimensions related to the generation, assembly, description and maintenance of data---reliability (with several sub-dimensions), metadata, security and confidentiality.\footnote{Modern industrial thinking about quality is, of course, dominated by focus on process rather than product \citep{deming-1982}. Interestingly, and  pertinent to this paper, the pioneer W. Edwards Deming spent a significant portion of his career at the U. S. Census Bureau. The increasingly important issue of data provenance also falls under process.}
\item[Data]
Dimensions specifically associated with the data themselves. At the record/table level, these comprise accuracy, completeness, consistency and validity. Database-level dimensions are identifiability and joinability.
\item[User]
Dimensions related to users and user---accessibility, integrability, interpretability, rectifiability, relevance and timeliness.
\item[Objectivity]
Dimensions describing whether disseminated information is accurate, reliable, scientifically sounds and unbiased in terms of both substance and presentation.
\item[Utility]
Dimensions addressing usefulness of the information for the intended audience’s anticipated purposes.
\item[Integrity]
Dimensions pertaining to protection of information from unauthorized, unanticipated or unintentional falsification or corruption.
\end{description}

Data quality cannot be disconnected from economics, because of the necessity to ask the question ``To what end have the data been collected?'' \citep{english99, karr-josdiscussion-2013}. Data quality-associated costs are imposed on multiple classes of stakeholders, among them, data subjects, data collectors and stewards, decision makers and society at large. There are both actual and opportunity costs. Whether incurring them is justified in a particular cases depends on the associated decisions.

Because of the large, public financial consequences, the field in which data quality has received the most attention and deepest investigation is official statistics. The entire field of total survey error (TSE) has emerged in response \citep{biemerlyberg03, groves-2004, tse-2017}, which is dominated by ``total quality'' thinking.

Measurement of data quality has long been a perplexing issue if construed narrowly to mean error rates, because ground truth is unknown. In \S\ref{sec.outliers} and \cite{markovstructure-2021}, we applied clustering to identify outliers that may be---and in synthesized cases, \emph{are}---data quality problems. In other settings---notably, statistical disclosure limitation, where data are altered deliberately in order to protect confidentiality, quality is measured by analytical utility. While sometimes problematic, because measures based on utility are often either too blunt or too narrow \citep{isr-2011}, this approach has been broadly productive \citep{framework-tas-2006, verificatioserverss-csda-2009}.

Uncertainty has not dominated data quality strategies, but has lurked in the background for years. Recent approaches such as ``fitness for purpose'' \citep{dg-fitness-2017} and the ``decision quality rather than data quality'' focus of \cite{karr-josdiscussion-2013} account implicitly for uncertainty. Re-formulated as ``Total Survey Uncertainty,'' \TSE\ can address uncertainty. Yet another means of accommodating uncertainties is explicit consideration of risk \citep{eltinge-biemer-holmberg-2013}.

\bibliographystyle{apalike}
\bibliography{e:/LaTeX/AFK}

\def\thisfile{TimesTemplate.tex}
\def\thisfiledate{2014/10/21}
\typeout{***** `\thisfile' <\thisfiledate> *****}

\end{document}